\documentclass{article}
\usepackage{arxiv}

\usepackage{lineno,epstopdf}
\usepackage{amsmath,amssymb,amsfonts}
\usepackage{multirow}
\usepackage{adjustbox}
\usepackage{graphicx}
\usepackage{textcomp}
\usepackage{color}
\usepackage{algorithm2e}
\RequirePackage[T1]{fontenc}
\RequirePackage[utf8]{inputenc}
\usepackage{array}
\newcolumntype{P}[1]{>{\centering\arraybackslash}p{#1}}

\SetCommentSty{mycommfont}
\modulolinenumbers[5]

\title{A scalable saliency-based Feature selection method with instance level information}

\author{
  Brais Cancela\thanks{Corresponding author.} \\
  CITIC. Universidade da Coruña.  \\
  15006, A Coruña, Spain. \\
  \texttt{brais.cancela@udc.es} \\
   \And
  Verónica Bolón-Canedo \\
  CITIC. Universidade da Coruña.  \\
  15006, A Coruña, Spain. \\
  \texttt{veronica.bolon@udc.es} \\
   \And
  Amparo Alonso-Betanzos \\
  CITIC. Universidade da Coruña.  \\
  15006, A Coruña, Spain. \\
  \texttt{amparo.alonso.betanzos@udc.es} \\
   \And
  Jo\~ao Gama \\
  LIAAD, INESCTEC. \\
  Rua Dr. Roberto Frias, 4200-465 Porto, Portugal. \\
  \texttt{jgama@fep.up.pt} \\
}

\begin{document}

\maketitle

\begin{abstract}
Classic feature selection techniques remove those features that are either irrelevant or redundant, achieving a subset of relevant features that help to provide a better knowledge extraction. This allows the creation of compact models that are easier to interpret. Most of these techniques work over the whole dataset, but they are unable to provide the user with successful information when only instance information is needed. In short, given any example, classic feature selection algorithms do not give any information about which the most relevant information is, regarding this sample. This work aims to overcome this handicap by developing a novel feature selection method, called Saliency-based Feature Selection (SFS), based in deep-learning saliency techniques. Our experimental results will prove that this algorithm can be successfully used not only in Neural Networks, but also under any given architecture trained by using Gradient Descent techniques.
\end{abstract}

\section{Introduction}

\begin{figure}[htp]
	\centering
	\begin{minipage}[]{0.95\linewidth}
	    \centering
 		\includegraphics[width=0.70\textwidth]{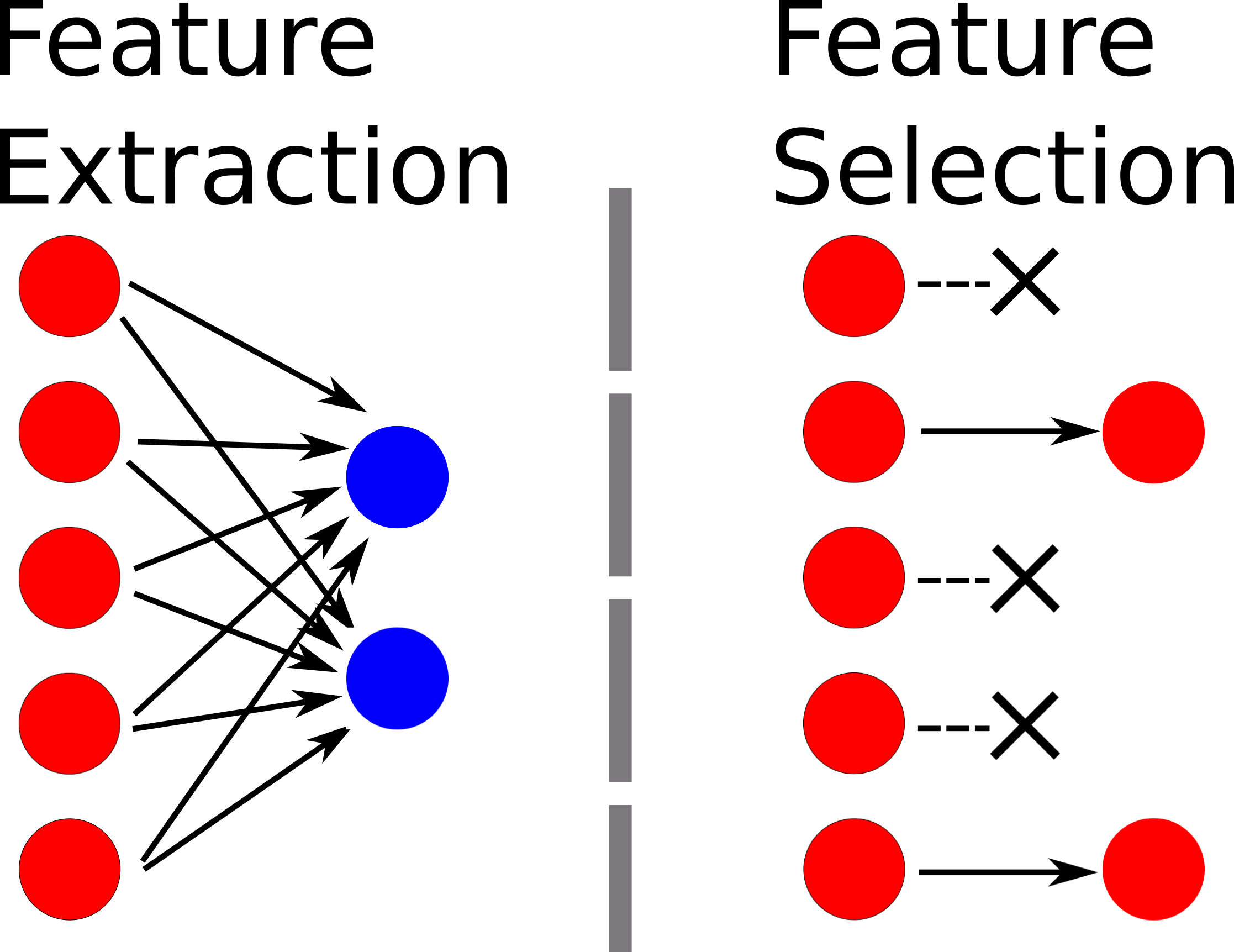}\\
	\end{minipage}
\caption{
 Feature Extraction (FE) creates new features by combining elements of the original input, whereas Feature Selection (FS) removes those features that are considered either irrelevant or redundant, while the rest are kept unaltered.
}
\label{fig:fs_vs_fe}
\end{figure}

With the rise of the so-called \textit{Big Data}, there is an increasing need for the use of techniques that allow us to reduce the input space \cite{hashem2015rise}. These techniques are often divided into two broad groups \cite{abe2010feature}: \textit{Feature Selection (FS)} and \textit{Feature Extraction (FE)}. Fig. \ref{fig:fs_vs_fe} shows a graphic representation about how these two approaches work.

On the one hand, Feature Extraction approaches reduce the number of characteristics by the combination (either linear or not) of the input space features \cite{guyon2006introduction}. For instance, a Feature Extraction technique in \textit{Deep Learning} is the so-called \textit{deep features}, which are the data representation that can be obtained if we remove the last Neural Network (NN) layer. Thus, these techniques are able to create a new feature set, which is often more compact and with higher discrimination capacity. This is the most used technique in image analysis, signal processing and information retrieval \cite{romero2016unsupervised,chen2016deep,krishnan2016deep}.

On the other hand, Feature Selection approaches achieve the dimensionality reduction by removing either irrelevant and redundant features \cite{guyon2003introduction}. Due to the fact that these techniques are able to preserve the \textit{original} features, it is an useful approach specially in applications where these attributes are essential to both understanding the model and for the knowledge inferring \cite{zou2015deep,han2015semisupervised,novakovic2016toward}. 

Often, FS techniques are divided in three big groups: \textit{filters}, that work independently of the inductive model (can be viewed as a data pre-processing step); and \textit{wrappers} and \textit{embedded methods}, which measure the feature relevance according to a classifiers' performance. However, the idea is quite similar: having any given dataset, the task is to select which are the features that contains the most discriminative information, in a global way. However, we aim to create an algorithm that, instead of using dataset-level information, takes sample-level information to build a FS model.


For example: suppose we have a medical dataset, and we aim to predict if a patient is prone to have cancer. Classic Feature Selection algorithms select which are the most important features that help the classifier to make a good prediction. On the contrary, we propose to infer, for any specific patient, which are the features the classifier is using to predict a certain output, building a FS algorithm by using this information. We believe this approach can keep all Classic Feature Selection advantages, while adding a new powerful tool: the ability to provide customized feature relevance information for each sample.

The most important work in model explainability is the LIME algorithm \cite{ribeiro2016should}. The idea is to apply small perturbations to the input, establishing the importance of each feature in the model's decision. Although this is a black box method that can be used in any given classifier model, it has two main drawbacks: 1) it is slow, as it requires too much time to evaluate each sample (up to 10 minutos in the ImageNet dataset, as the authors have mentioned in their work); and 2) it cannot help the model to improve its explainability, as it is designed to be used after the model is properly trained.

In this work we propose to solve this problem by introducing a novel Feature Selection algorithm, called Saliency-based Feature Selection (SFS). Our proposal consists in using techniques that are able to provide personalized information for any given example. Once this information is collected, a Feature Selection algorithm is created by including those features that contain a higher discrimination coefficient. As at present, scalability is also an important requirement in Machine Learning algorithms, our algorithm must fulfill also another specific requisite: it must be able to work under \textit{Big Data} environments. 

The rest of the paper is organized as follows: section \ref{sec:saliency} will explain the personalized information algorithm we are going to use; section \ref{sec:gain_function} will provide the metrics used to compute our saliency; section \ref{sec:saliency_fs} will propose our novel SFS algorithm; section \ref{sec:ablation} will show an ablation study regarding our approach; finally, section \ref{sec:results} will show some experimental results over public datasets, and section \ref{sec:conclusion} will offer some conclusions and future work.

\section{Personalized Information Algorithm}
\label{sec:saliency}

There are only a few research works that address the problem of FS in Big Data environments.  In \cite{ibrahim2014multi, zou2015deep} the authors use Deep Bayesian Networks (Deep BN) to select the most relevant features. Although Deep BN can be used in Big Data, they only provide results in small datasets (either few examples or few features). In \cite{feng2017data}, the authors claimed to use a deep feature selection technique to reduce the input space in short-term wind forecasting models. However, their approach uses Recursive Feature Elimination (RFE) which requires to exponentially train several different models, making it unable to be used with a high number of features.

Perhaps the most interesting work is presented in \cite{li2016deep}. It is called Deep Feature Selection (DFS), and consists on a Elastic Net variant, that can be introduced as an extra layer into any Neural Network (NN). However, the authors claim that the number of layers has to be reduced so as to have the  method properly working. Thus, it is not suitable for working with Convolutional Neural Networks (CNN) models. This method introduces a mask in the input data, adding a $l_1$ and $l_2$-regularization to that mask, in the same way the Elastic Net is defined. Furthermore, Elastic Net penalties are also applied to the hidden layers.
 
However, as early mentioned, none of these approaches is able to, for any given example, give personalized information. To address this issue, we propose to use a well-known computer vision technique, which is called \textit{saliency}.

\subsection{Saliency}

\textit{Saliency} is a technique that was first developed in computer vision problems. Its goal is to evaluate the \textit{degree of quality} for each pixel within an image \cite{simonyan2013deep}. Often, NNs are seen like a black box where, given any input and any desired output, it is possible to obtain an accurate prediction that is somehow close to what we should expect. However, NNs do not provide any kind of transparent explanation about how the system reaches the predicted solution. Saliency was created with the purpose of \textit{seeing} what is happening inside the NN. Nowadays, there are two different approximations to calculate this saliency: supervised or semi-supervised.
 
The first works used a semi-supervised approach \cite{simonyan2013deep,mahendran2015understanding}. The idea is simple: having any given trained NN for classification, and any given image, a back-propagation routine is used to detect which are the pixels that most influenced the desired output.

Supervised approaches are more recent \cite{zhao2015saliency,zhang2017co}, and they are also trained in a different way. In this case, the NN has the same input and output size. Furthermore, we also know, for any given image in the training set, which are the most relevant features. For instance, if we are detecting a cat, the important features are the pixels in the image that belong to the cat. The model is trained so that the predicted output matches our previous segmentation of important features. This is the reason why these networks are also called \textit{Semantic Segmentation Networks}. They are also called \textit{Attention Models} \cite{mnih2014recurrent,xu2015show} whenever a Recurrent Neural Network is used at the end to evaluate the quality of the feature.

Supervised techniques achieve better results than semi-supervised, but they have a major drawback: it is necessary to know, a priori, which are the most relevant features for each instance, in order to successfully train the model. In the case of an image dataset it is easy, but it is not always possible to obtain this information in other environments, like in the case of DNA microarrays, for instance. Furthermore, the supervised techniques can only be used in classification problems. They are not suitable for other approaches, like regression.

Because of that, we propose to use a semi-supervised saliency technique.

\subsection{Our Approach}

For our model, we are going to use a generalization of the idea proposed in \cite{simonyan2013deep}. Let $\mathbf{X} \in \mathcal{R}^{N \times R}$ be our input data, with $N$ being the number of samples and $R$ the number of features; and let $\mathbf{\tilde{Y}} = f(\mathbf{X}; \mathbf{\Theta}) \in \mathcal{R}^{N \times C}$ be our classification model (just for the purposes of explaining the model; later we will show how our approach can also be applied in regression problems). It does not matter which type of model we are employing, as long as it can be trained by using a loss minimization function (NN, CNN, SVM, \ldots). $C$ is the number of different classes to evaluate, and $\mathbf{\Theta}$ are the classifier weights, which are adjusted during the training procedure.

For the purpose of explanining the idea, we can assume that $f(\mathbf{X}; \mathbf{\Theta})$ is the result of applying the \textit{softmax} function to a one layer model ($\Theta \in \mathcal{R}^{C \times R}$). Thus, we have that
\begin{equation}
    \sum_{c = 1}^C y_c^{(i)} = 1,
\end{equation}
where
\begin{equation}
    y_c^{(i)} = \mbox{softmax}(\theta^T_c \mathbf{X}^{(i)}).
\end{equation}

$y_c^{(i)}$ is the probability that the instance $i$ belongs to the class $c$. $\theta_c$ is the $c$-th column of $\Theta$.

To train this model, a loss function $\ell(\mathbf{\Theta}; f, \mathbf{X}, \mathbf{Y})$ is minimized, where $\mathbf{Y} \in \mathcal{R}^{N \times C}$ is the class one-hot encoding. Since we are using the softmax function as our output, our minimization function is the \textit{categorical cross-entropy}, defined as 
\begin{equation}
    \label{eq:cross_entropy}
    \ell(\mathbf{\Theta}; f, \mathbf{X}, \mathbf{Y}) = - \dfrac{1}{N} \sum_{i=1}^N \sum_{c=1}^C y^{(i)}_c \log \left(f(\mathbf{x}^{(i)}; \mathbf{\Theta})_c\right)
\end{equation}

\subsubsection{Classic Saliency} \quad In order to know which the features most contribute to activating the class $c$, the solution proposed in \cite{simonyan2013deep} is to evaluate the gradient of $y_c^{(i)}$ with respect to the input. To put it in mathematical terms,
\begin{equation}
    \sigma^{(i)}_c = \bigg | \dfrac{\partial y_c^{(i)}}{\partial \mathbf{X}^{(i)} } \bigg |.
\end{equation}

To give an intuition behind the idea, this gradient indicates how we should modify the input instance in order to maximize its belonging to class $c$. This technique, however,  only works in classification problems. Thus, we need to create a generalization of this method for our purposes.

\subsubsection{Generalization Approach} \quad Instead of applying a gradient function for each class $c$, our idea is similar to the one that is used to update the weights during the training procedure. As the loss function usually gives higher gradients whenever there is a strong training misclassification, we propose the definition of a \textit{Gain Function} ($g(\mathbf{\Theta}; f, \mathbf{X}, \mathbf{Y})$), that will bring high gradients whenever a sample is correctly classified. Thus, our \textit{Saliency Function} $\sigma$ will be defined as
\begin{equation}
    \sigma(\tilde{y}^{(i)}, y^{(i)}) = \left| \dfrac{\partial \text{\textit{g}}(\tilde{y}^{(i)})}{\partial \mathbf{x}^{(i)}} \right|,
\end{equation}
where $\mathbf{\tilde{y}} = f(\mathbf{X}; \mathbf{\Theta})$ is our model's predicted output for the instance $i$. 

Below we will show how to create this \textit{Gain Function} $g$, depending on the loss function we are trying to minimize.

\vspace{1cm}

\section{Building the Gain Function}
\label{sec:gain_function}

As early mentioned, our aim is to develop a saliency system that can work with multiple types of problems. To that end, we are going to explain how to create our Gain Function $g$ in three different scenarios: one for regression and two for classification (NNs and SVMs).

\subsection{Regression}

First, we will introduce the Regression Gain Function, as it is the most intuitive one. As simplification, we assume our model is trained by using the Mean Square Error Loss (MSE):
\begin{equation}
    \ell_{MSE}(\mathbf{\tilde{Y}}, \mathbf{Y}) = - \dfrac{1}{N} \sum_{i=1}^N \left( \mathbf{\tilde{y}}^{(i)} - \mathbf{y}^{(i)} \right)^2,
\end{equation}
where $\mathbf{\tilde{Y}} = f(\mathbf{X}; \mathbf{\Theta})$ is our model's predicted output. It does not matter if you choose a different loss function, because all regression losses have the same structure: the loss is $0$ if the prediction is perfect, and the loss increases as the prediction moves away from the expected result.

On the contrary, our Gain Function must behave in the opposite way: it must have high values whenever the prediction is good, and values close to zero with poor predictions. Thus, our solution is to use the inverse of the MSE loss function:
\begin{equation}
    \label{eq:mse_gain_function}
    g_{MSE}(\mathbf{\tilde{Y}}, \mathbf{Y}) = \dfrac{\alpha}{\ell_{MSE}(\mathbf{\tilde{Y}}, \mathbf{Y}), + \epsilon},
\end{equation}
where $\alpha$ is a multiplication factor and $\epsilon > 0$ is a factor to avoid division by zero. By default, we set $\alpha = 1$ and $\epsilon = 10^{-3}$.

\subsection{Classification}

Although Eq. \ref{eq:mse_gain_function} might also be used as the Gain Function in classification problems, we found it not very suitable. The reason is its behavior when a total misclassification ($\tilde{y}^{(i)}_c = 0$ and  $y^{(i)}_c = 1$, for instance) occurs.
Recall that we are using the saliency function to build a feature selection algorithm. Thus, we do not want to receive any information when a total misclassification occurs. That is, our Gain Function should return $0$ values when this happens. Unfortunately, the Gain Function provided in Eq. \ref{eq:mse_gain_function} does not satisfy this requirement.

Consequently, we have developed two different gain functions for two different classification losses: \textit{cross-entropy} and \textit{hinge} loss, the ones most commonly used. 

\subsubsection{Cross-entropy Gain Function} The cross-entropy loss (Eq. \ref{eq:cross_entropy}) is the most common used loss function when dealing with NN for Classification, including CNNs. Since our Gain Function should operate in the opposite way, as discussed above, our proposed solution is
\begin{equation}
    g_{CE}(\mathbf{\tilde{Y}}, \mathbf{Y}) = - \dfrac{\alpha}{N} \sum_{i=1}^N \sum_{c=1}^C y^{(i)}_c \log \left(1 - \hat{y}^{(i)}_c \right),
\end{equation}
where 
\begin{equation}
    \label{eq:clip}
    \hat{y}^{(i)}_c = \min \left\{1 - \epsilon, \quad \tilde{y}^{(i)}_c \right\},
\end{equation}
in order to avoid a zero-logarithm. $\alpha$ and $\epsilon$ have the same behavior exposed in Eq. \ref{eq:mse_gain_function}. This function will ensure that no saliency is obtained whenever a total misclassification occurs $(\log(1 - 0) = \log(1) = 0)$. Note that we do not want to kill the gradient when clipping the value in Eq. \ref{eq:clip}. Thus, the gradient should remain unaltered ($\nabla \hat{y}^{(i)}_c = 1$).

Note that this idea is somehow similar to the one proposed in \cite{simonyan2013deep}. It only differs in one point: while the method proposed in \cite{simonyan2013deep} decomposes the last layer network to obtain the saliency, we applied our model directly to a Gain Function, which makes it suitable to use it in other machine learning approaches, as regression. Another advantage of our approach is that it returns close-to-zero gradients whenever there is a misclasification. In this sense, our algorithm is able to indicate that there are no relevant features in a misclassification. On the contrary, the saliency defined in \cite{simonyan2013deep} does not take into account this crucial information.

\subsubsection{Hinge Gain Function} \quad The hinge loss is often used to train SVMs. In a multiclass problem, it can be defined as 
\begin{equation}
    \label{eq:hinge}
    \ell_H(\mathbf{\tilde{Y}}, \mathbf{Y}) = - \dfrac{1}{N} \sum_{i=1}^N \sum_{c=1}^C y^{(i)}_c \max(0, 1 - \tilde{y}^{(i)}_c) +
    (1 - y^{(i)}_c) \max(0, 1 + \tilde{y}^{(i)}_c),
\end{equation}
that is, our correct class prediction will have values higher than $1$, whereas values lower than $-1$ will be obtained for the incorrect classes, as desired.

We may note that the function does not have gradient when the values are higher than $1$ in the correct class (same with $-1$ in the incorrect ones). Thus, a predicted output with value of $2$ should have the same information as a predicted value of $2000$. Thus, we modified the Gain Function as follows:
\begin{equation}
    g_{H}(\mathbf{\tilde{Y}}, \mathbf{Y}) = - \dfrac{\alpha}{N} \sum_{i=1}^N \sum_{c=1}^C y^{(i)}_c \log \left(1 - \Breve{y}^{(i)}_c \right),
\end{equation}
where 
\begin{equation}
    \Breve{y}^{(i)}_c = \min \left\{1 - \epsilon, \quad \dfrac{\min(1, \max(-1, \tilde{y}^{(i)}_c)) + 1}{2} \right\}.
\end{equation}

Again, we do not kill the gradient after clipping ($\nabla \Breve{y}^{(i)}_c = 1$).


\section{Saliency-based Feature Selection}
\label{sec:saliency_fs}

Our approach is named as Saliency-based Feature Selection, or SFS. It is a ranker-based feature selection method, that is, it returns an ordered vector of all features, based on their importance. In Algorithm \ref{alg:saliency_fs} we show its schema.

\begin{center}
\begin{minipage}{.55\linewidth} 
\begin{algorithm}[H]
 \KwData{$\mathbf{X}, \mathbf{Y}, f, \ell, \mathbf{\Theta}, \gamma, \epsilon, \mbox{reps}$}
 \KwResult{feature ranking $\mathbf{r}$}
 $n_f \leftarrow R$ \quad \tcp{$n_f$ is the number of \emph{Alive} features} 
 $\mathbf{r} \leftarrow [1 \ldots n_f]$\;
 \While{$n_f > \epsilon > 1$}{
    $\mathbf{\hat{X}} \leftarrow \mathbf{X}$\;
    $\mathbf{\hat{X}}[:, \mathbf{r}[n_f+1:R]] \leftarrow 0$\;
    $\mathbf{\sigma}_{fs} \leftarrow \mbox{zeros}(n_f)$\;
    \For{$\mbox{rep}\gets1$ \KwTo $C$}{
        Initialize $f(\mathbf{\hat{X}}; \mathbf{\Theta})$\;
        Train $f(\mathbf{\hat{X}}; \mathbf{\Theta})$ given $\mathbf{Y}$\;
        $\mathbf{\tilde{Y}} \leftarrow f(\mathbf{\hat{X}}; \mathbf{\Theta})$\;
        $\mathbf{\sigma}_{fs} \leftarrow \mathbf{\sigma}_{fs} + \mbox{GetSaliency}(\tilde{Y}, Y, \sigma)$\;
    }
    $\mbox{index} \leftarrow \mbox{argsort}(\mathbf{\sigma}_{fs}, \mbox{descend})$\;
    $\mathbf{r}[1:n_f] \leftarrow \mathbf{r}[\mbox{index}]$\;
    $n_f \leftarrow int(n_f * \gamma)$\;
 }
 \caption{Pesudocode of the SFS Algorithm}
 \label{alg:saliency_fs}
\end{algorithm}
\end{minipage}
\end{center}

The proposed algorithm only contains three hyper-parameters: $\epsilon \geq 1$, a stopping criteria variable; $1 \geq \gamma > 0$, which controls the number of \textit{alive} features that are kept in the next iteration; and $reps$, which controls the number of times a model is trained, in order to avoid overfitting.

We start by training the model $f$ with all the features in the feature set. Then, we train the model and we compute the saliency. After that, we sum up and sort all features, obtaining the feature ranking $\mathbf{r}$. Then, we discard the least relevant features, and we repeat the operation until the stopping criteria is satisfied.

\begin{center}
\begin{minipage}{.55\linewidth} 
\begin{algorithm}[H]
    \SetKwFunction{FMain}{GetSaliency}
    \SetKwProg{Fn}{Function}{:}{}
    \Fn{\FMain{$\tilde{Y}, Y, \sigma$}}{
        $\mathbf{\sigma} \leftarrow 0$\;
        $C \leftarrow $ Number of classes in $Y$\;
        \For{$c\gets1$ \KwTo $C$}{
            $\mathbf{\sigma}_c \leftarrow \sum_{i_c=1}^{N_c} \sigma(\mathbf{\tilde{y}}^{(i_c)}, \mathbf{y}^{(i_c)})$\;
            $\mathbf{\sigma} \leftarrow \mathbf{\sigma} + \dfrac{\mathbf{\sigma}_c}{\| \mathbf{\sigma}_c \|_1}$\;
        }
        \textbf{return} $ \mathbf{\sigma} $ 
    }
    \vspace{0.1cm}
    \textbf{end}
    \vspace{0.2cm}
 \caption{Saliency Function for Classification}
 \label{alg:saliency_fs_classification}
\end{algorithm}
\end{minipage}
\end{center}

The way to compute the saliency differs depending on whether we are dealing with a classification or a regression problem. In the case of the classification issue, the procedure is explained in Algorithm \ref{alg:saliency_fs_classification}. Basically, we compute and normalize the saliency for each class. Alter that, we sum up all features. In case of a regression problem, we just sum up all the saliency scores, as described in Algorithm \ref{alg:saliency_fs_regression}.

\begin{center}
\begin{minipage}{.55\linewidth} 
\begin{algorithm}[H]
    \SetKwFunction{FMain}{GetSaliency}
    \SetKwProg{Fn}{Function}{:}{}
    \Fn{\FMain{$\tilde{Y}, Y, \sigma$}}{
        $\mathbf{\sigma} \leftarrow \sum_{i=1}^{N} \sigma(\mathbf{\tilde{y}}^{(i)}, \mathbf{y}^{(i)})$\;
        \textbf{return} $ \mathbf{\sigma} $ 
    }
    \vspace{0.1cm}
    \textbf{end}
    \vspace{0.2cm}
 \caption{Saliency Function for Regression}
 \label{alg:saliency_fs_regression}
\end{algorithm}
\end{minipage}
\end{center}

The complexity of this algorithm is variable, as it completely depends on the $\gamma$ parameter. In the best scenario, when $\gamma = 0$, the complexity is lineal in the number of instances ($\mathcal{O}(N)$), whereas in the worst scenario $\gamma \approx 1$, the complexity also depends on the number of variables ($\mathcal{O}(RN)$), as we barely remove one feature in each loop.

\section{Ablation Study}
\label{sec:ablation}

In this section we will show how parameters $\gamma$ and \textit{reps} affect the behavior or our algorithm. Furthermore, we will show how decoupling the model to obtain the feature ranking from the model that is used to classify can affect our approach. In order to accomplish that, we first introduce the datasets that will be used to test our methodology.

\subsection{Datasets}

\begin{table*}
	\centering
	\caption{NIPS 2003 Feature Selection Challenge datasets.}
	\label{tab:datasets}
	\begin{adjustbox}{max width=\columnwidth}
	\begin{tabular}{| c | c | c | c | c | c |} 
      	\hline
      	  Name & \# instances (train, test) & \# features & \# relevant features & \% relevant features & pos/neg ratio \\
      	\hline 
        Arcene & (88, 112) & 10000 & 7000 & 0.7 & 1.0\\
      	\hline
        Dexter & (300, 300) & 20000 & 9947 & 0.5 & 1.0\\
      	\hline 
        Dorothea & (800, 350) & 100000 & 50000 & 0.5 & 0.11\\
      	\hline 
        Gisette & (6000, 1000) & 5000 & 2500 & 0.5 & 1.0\\
      	\hline 
        Madelon & (2000, 600) & 500 & 20 & 0.04 & 1.0\\
      	\hline 
    \end{tabular}
    \end{adjustbox}
\end{table*}

\subsubsection{NIPS 2003 Feature Selection Challenge}
We have selected the 5 datasets that were proposed in the \textit{NIPS 2003 Feature Selection challenge}\footnote{http://clopinet.com/isabelle/Projects/NIPS2003/}. There are 5 synthetic datasets that were designed with the only purpose of measuring the quality of feature selection algorithms for classification. The specific characteristics of each dataset are shown in Table \ref{tab:datasets}. Although each dataset only contains two classes, they are challenging because of these factors: few training examples (Arcene), unbalanced data (Dorothea) or low relevant features (Madelon).

\subsubsection{MNIST, FASHION-MNIST, CIFAR-10 and CIFAR-100}

Due to their small number of examples, NIPS 2003 FS challenge datasets are not suitable to test the behavior of our approach in Big Data scenarios. To overcome this issue, we have included four classic computer vision datasets:
\begin{itemize}
\item {MNIST} It is the most classic computer vision classification challenge \cite{lecun1998gradient}. It contains more than 50.000 handwritten digits (10 classes) stored in grayscale $28 \times 28$ images.
\item {FASHION-MNIST} It is a dataset of Zalando's article images \cite{xiao2017_online}. It contains 60.000 images (10 classes) stored in grayscale $28 \times 28$ resolution.
\item {CIFAR-10 and CIFAR-100} They were proposed in \cite{krizhevsky2009learning}. Each one contains more than 50.000 tiny RGB images ($32 \times 32 \times 3$) of different objects (car, truck, plane, \ldots). The first one contains images belonging to 10 different labels, whereas the other contains images from 100 different classes.
\end{itemize}
\subsubsection{Regression}

As early mentioned, and different from classic information-based FS algorithms \cite{bolon2013review}, our model is able to perform FS in regression problems. To test it we have used two different Big Data datasets: 

\begin{itemize}
    \item \textit{Relative location of CT slices on axial axis\footnote{https://archive.ics.uci.edu/ml/datasets/Relative+location+of+CT+slices+on+axial+axis}}. Originally published in \cite{graf20112d}, its aim is to discover the relative location of the image on the axial axis. It contains 53500 CT images that belong to 74 different patients. Each image is reduced to two different histograms, resulting in 385 total features.
    \item \textit{Enery Molecule Dataset\footnote{https://www.kaggle.com/burakhmmtgl/energy-molecule}}. Originally published in \cite{himmetoglu2016tree}, the dataset contains the ground state energies of 16,242 molecules, each with 1275 features. The aim of this dataset is to use Machine Learning techniques to quickly compute the atomization energy, as the simulations needed to compute it require a high computational time.
\end{itemize}

\subsection{Testing the effects of the algorithm´s parameters}
\subsubsection{Effect of `reps' parameter}
\label{sec:reps}
\begin{figure*}
	\centering
	\begin{minipage}[]{0.49\linewidth}
	    \centering
 		\includegraphics[width=0.95\textwidth]{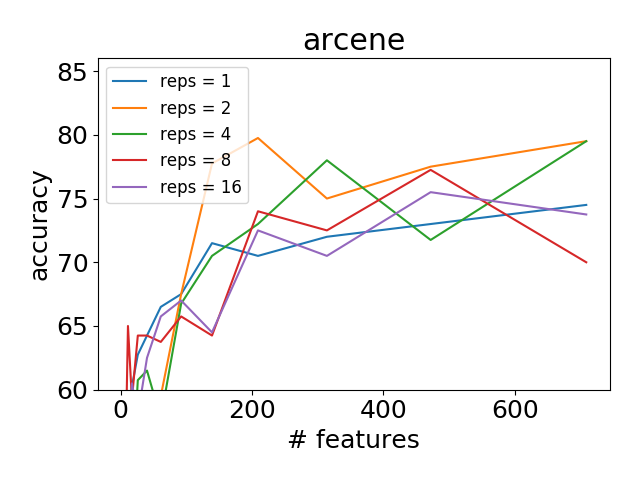}\\
	\end{minipage}
	\centering
	\begin{minipage}[]{0.49\linewidth}
	    \centering
 		\includegraphics[width=0.95\textwidth]{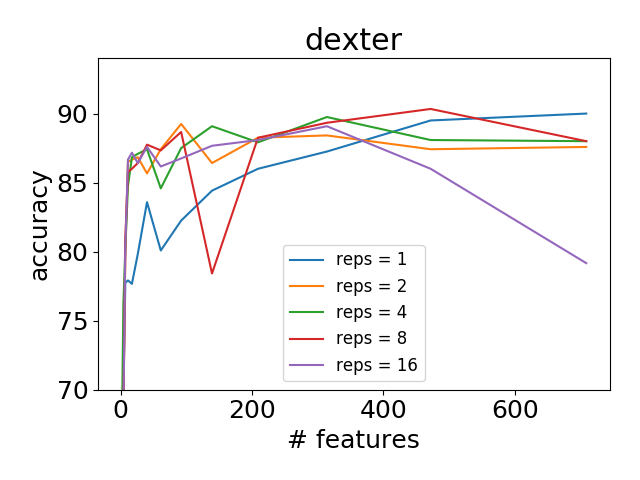}\\
	\end{minipage}
	\centering
	\begin{minipage}[]{0.49\linewidth}
	    \centering
 		\includegraphics[width=0.95\textwidth]{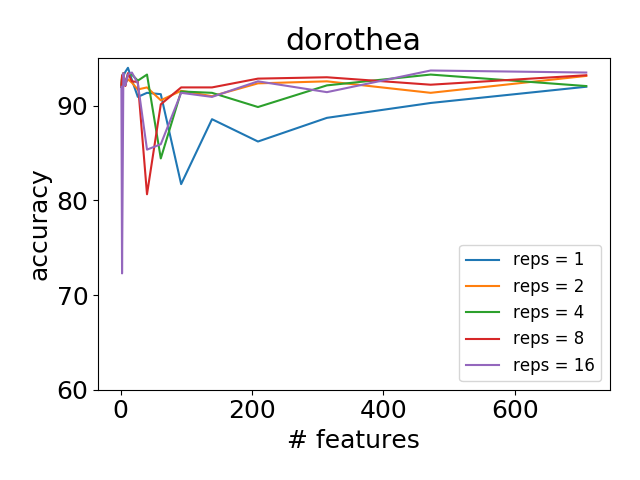}\\
	\end{minipage}
	\centering
	\begin{minipage}[]{0.49\linewidth}
	    \centering
 		\includegraphics[width=0.95\textwidth]{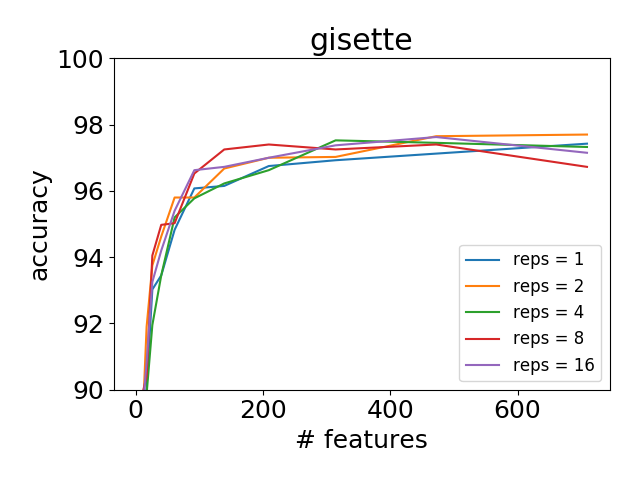}\\
	\end{minipage}
	\centering
	\begin{minipage}[]{0.49\linewidth}
	    \centering
 		\includegraphics[width=0.95\textwidth]{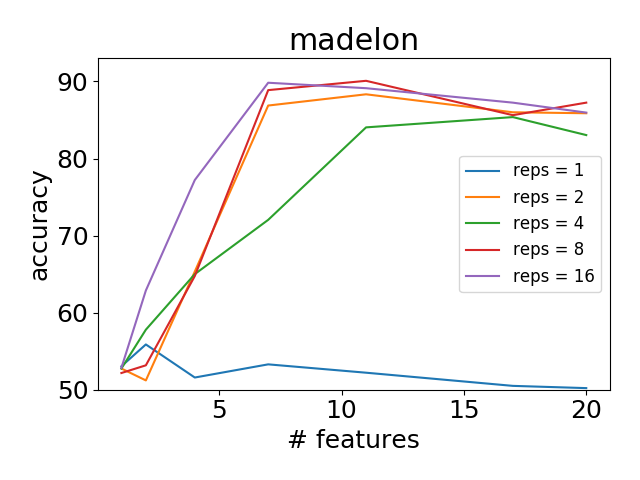}\\
	\end{minipage}
\caption{
 Effect of the number of training repetitions over our algorithm. The number of repetitions significantly affects the obtained result.
}
\label{fig:reps}
\end{figure*}

As it is well-known, it is almost impossible to train the same machine learning model more than once, expecting to achieve the very same exact result each time. Aspects like random initialization in model's weights, or random permutations in the training set cause the model output to be similar but not entirely exact each time the model is trained. Thus, our first aim was to evaluate how the number of train repetitions can affect the output. Our objective was to check if it will be necessary to, at each step, train the model more than once, computing the saliency ranking as the mean average of all repetitions.

We have trained our algorithm using the NIPS 2003 FS Challenge datasets, fixing $\gamma = 0$, and trying different number of repetitions. We have trained a 3-layer Fully-Connected NN ($150$, $100$ and $50$ nodes per layer, respectively). Batch Normalization (BN) \cite{ioffe2015batch} and $\text{\textit{ReLU}}(x) = max(0, x)$ activation are used. Softmax function is used as output, and Eq. \ref{eq:cross_entropy} is used as loss function.

We have included an l2 \textit{weight decay} regularization with factor $0.001$. We have trained the model for 100 epochs, using Adam \cite{kingma2014adam} as optimizer. To deal with unbalanced data, we have replicated the number of examples until we achieved a true balance between classes. All models were created using Keras Framework\footnote{https://keras.io/}, along with TensorFlow \cite{abadi2016tensorflow} as back-end.

Fig. \ref{fig:reps} shows the results obtained. It can be seen that the number of repetitions used affects accuracy in most datasets, and that this effect depends also in the number of features. We have conducted a Friedman test, resulting in there is no significant difference between the models when the number of repetitions is higher than 2, except in the Madelon dataset, where there is a low number of relevant features. We also found significant differences between these models against the model with just one repetition. Thus, we recommend to use a number of repetitions higher than 2 to achieve a better performance.

\subsubsection{Effect of $\gamma$ parameter}
\begin{figure*}
	\centering
	\begin{minipage}[]{0.49\linewidth}
	    \centering
 		\includegraphics[width=0.95\textwidth]{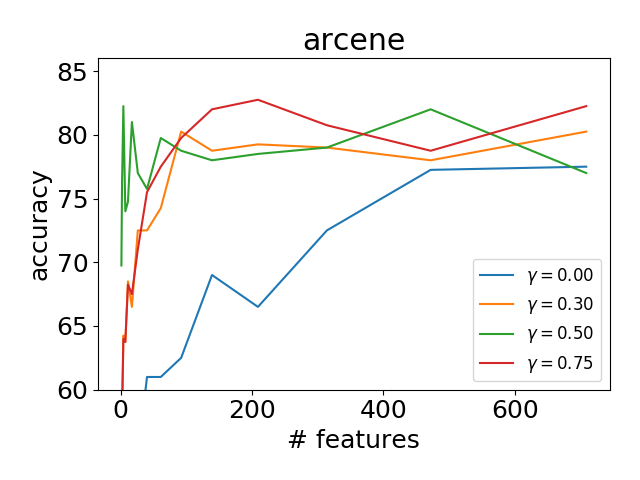}\\
	\end{minipage}
	\centering
	\begin{minipage}[]{0.49\linewidth}
	    \centering
 		\includegraphics[width=0.95\textwidth]{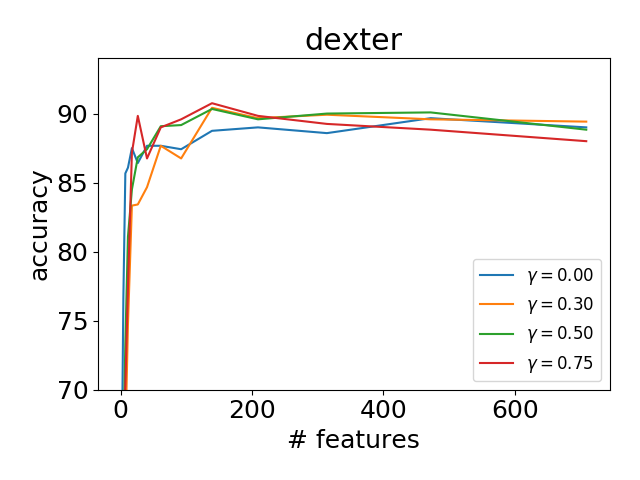}\\
	\end{minipage}
	\centering
	\begin{minipage}[]{0.49\linewidth}
	    \centering
 		\includegraphics[width=0.95\textwidth]{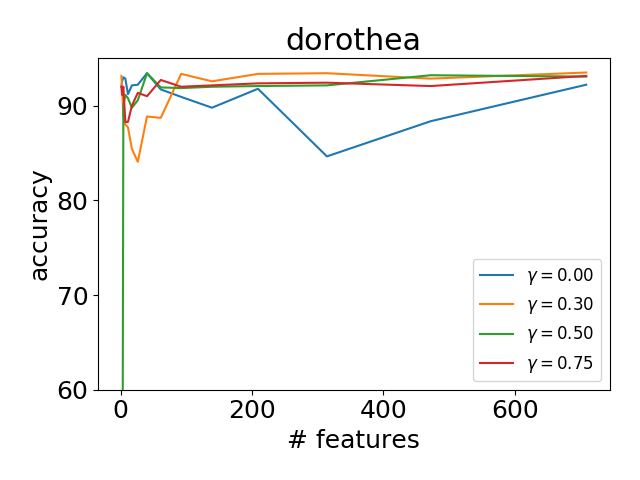}\\
	\end{minipage}
	\centering
	\begin{minipage}[]{0.49\linewidth}
	    \centering
 		\includegraphics[width=0.95\textwidth]{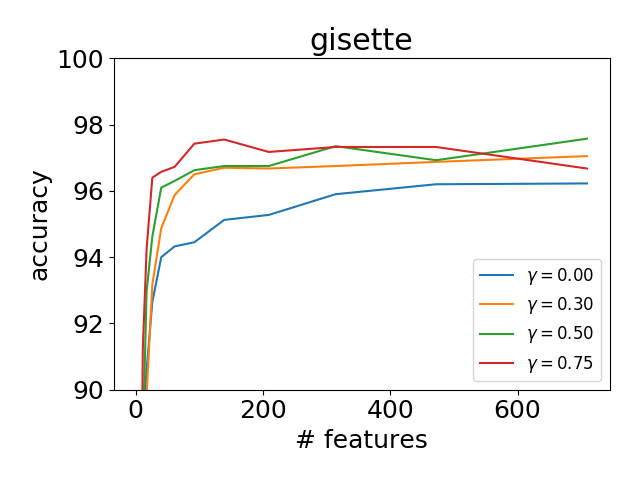}\\
	\end{minipage}
	\centering
	\begin{minipage}[]{0.49\linewidth}
	    \centering
 		\includegraphics[width=0.95\textwidth]{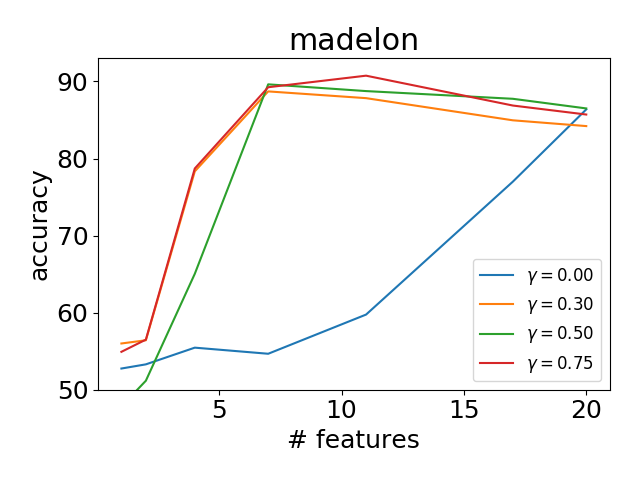}\\
	\end{minipage}
\caption{
 Effect of the hyper-parameter $\gamma$ over our algorithm. Using $\gamma = 0$ led the system to achieve poor results when using a low number of features. However, there is no significant performance improvement when $\gamma > .3$. 
}
\label{fig:gamma}
\end{figure*}

In order to carry out the experiments to analyze the case of the behavior of the$\gamma$ parameter, we have fixed $\mbox{reps} = 1$, while rest of parameters were the same as the in previous subsection. Fig. \ref{fig:gamma} shows that the accuracy of our algorithm improves as the $\gamma$ value increases. This occurs because when removing some features, some redundancy is also eliminated, causing some other features to become more important. Although increasing the $\gamma$ value helps our algorithm to achieve better scores, the Friedman tests we have conducted suggest there are no significant differences when we select $\gamma \geq 0.3$, except in the Madelon dataset. Furthermore, a Wilcoxon test suggest there is a significant difference between $\gamma = 0$ and $\gamma = 0.75$ models over the datasets Arcene, Madelon and Gisette. However, we found no significant difference over the Dexter and Dorothea datasets. In these datasets, we presume the high features/samples ratio are causing this effect.

\subsubsection{Effect of decoupling ranker and classifier}

Our Feature Selection algorithm is an embedded model, as both selection and classification/regression tasks can be performed at the same time. An embedded system selects features that achieve good results in the same model that is used to perform either the classification or the regression task. However, this might led us to select features that are only valid for the machine learning model we are using to obtain the ranking.

\begin{table*}
	\centering
	\caption{Decoupling ranker and classifier accuracy results. In brackets, number of features used to achieve the best score. In black, the best score achieved by each classifier. If multiple rankers achieve the best score, only the one with fewer features is highlighted.}
	\label{tab:decoupling}
	\begin{adjustbox}{max width=\columnwidth}
	\begin{tabular}{| c | c | c | c | c | c |} 
      	\hline
      	  \multirow{2}{*}{Dataset} & \multirow{2}{*}{SVM Ranker} & \multicolumn{4}{|c|}{SVM Classifier (\# of features)}\\
      	\cline{3-6} 
      	  & & Linear & Poly & RBF & Sigmoid \\
      	\hline
        \multirow{4}{*}{Arcene} & Linear & 83.0 (3189) & \textbf{88.0} (9750) & 85.0 (48) & \textbf{75.0} (327) \\
      	\cline{2-6} 
         & Poly & 84.0 (5578) & \textbf{88.0} (9750) & 81.0 (1088) & 73.0 (438) \\
      	\cline{2-6} 
      	 & RBF & 83.0 (1269) & \textbf{88.0} (9750) & \textbf{88.0} (487) & 74.0 (3622) \\
      	\cline{2-6} 
      	 & Sigmoid & \textbf{84.0} (1145) & \textbf{88.0} (9750) & 81.0 (1088) & 72.0 (3715) \\
      	\hline
      	\hline
        \multirow{4}{*}{Dexter} & Linear & \textbf{94.3} (1419) & 93.3 (99) & \textbf{93.7} (105) & \textbf{93.0} (102) \\
      	\cline{2-6} 
         & Poly & 93.6 (7071) & 91.0 (66) & 90.7 (33) & 90.3 (68) \\
      	\cline{2-6} 
      	 & RBF & 94.0 (7071) & 92.0 (93) & 92.7 (37) & 91.7 (40) \\
      	\cline{2-6} 
      	 & Sigmoid & 93.7 (3384) & \textbf{93.7} (66) & 92.3 (54) & 92.0 (64) \\
      	\hline
      	\hline
        \multirow{4}{*}{Dorothea} & Linear & 94.9 (222) & 94.6 (216) & 94.6 (79) & \textbf{95.1} (11318) \\
      	\cline{2-6} 
         & Poly & \textbf{94.9} (69) & \textbf{95.1} (88) & \textbf{94.9} (65) & 94.9 (150) \\
      	\cline{2-6} 
      	 & RBF & 94.6 (26) & 94.6 (25) & 94.6 (23) & 94.9 (27) \\
      	\cline{2-6} 
      	 & Sigmoid & 94.3 (71) & 94.6 (75) & 94.9 (96) & 94.9 (85) \\
      	\hline
      	\hline
        \multirow{4}{*}{Gisette} & Linear & \textbf{98.3} (1080) & 98.3 (714) & 98.1 (351) & 97.6 (351) \\
      	\cline{2-6} 
         & Poly & 98.1 (168) & 98.2 (360) & 98.2 (470) & 97.9 (240) \\
      	\cline{2-6} 
      	 & RBF & 98.2 (275) & 98.2 (581) & 98.2 (412) & 98.0 (412) \\
      	\cline{2-6} 
      	 & Sigmoid & 98.0 (315) & \textbf{98.3} (483) & \textbf{98.3} (351) & \textbf{98.0} (351) \\
      	\hline
      	\hline
        \multirow{4}{*}{Madelon} & Linear & 58.2 (218) & 71.7 (94) & 73.7 (271) & \textbf{57.0} (374) \\
      	\cline{2-6} 
         & Poly & 61.0 (9) & \textbf{76.2} (18) & 88.0 (11) & 52.3 (500) \\
      	\cline{2-6} 
      	 & RBF & 60.5 (5) & 70.3 (96) & 89.3 (12) & 53.7 (3) \\
      	\cline{2-6} 
      	 & Sigmoid & \textbf{62.0} (4) & 71.3 (108) & \textbf{90.8} (17) & 52.3 (500) \\
      	\hline
    \end{tabular}
    \end{adjustbox}
\end{table*}

This fact arises one question: \textit{How good is our selection?}. Consequently, we tested our proposal separating the problem into two different tasks: ranking and classifying. By using different models for each task, we will test how dependant is our ranking with respect to the model that was used to obtain it. To this end, we used the four kernel implementations provided by the sklearn's SVC dataset \cite{scikit-learn}, fixing the parameters $C = 1$, $degree = 3$ and $coef0 = 1$. Our algorithm meta-parameters were fixed to $\gamma = 0.975$ and $reps = 1$. We do not need to use more repetitions, as the SVM SMO-training algorithm \cite{platt1998sequential} is very stable. Table \ref{tab:decoupling} shows the obtained results over the NIPS datasets.

Two main answers arise:
\begin{enumerate}
    \item Only in $35\%$ of cases the best result is provided by using the same algorithm to obtain both the ranking and the classification. Since we are using 4 different models, this percentage is close to random.
    \item In three different datasets (Arcene, Dorothea and Gisette) the best ranker is the same for three different classifiers. This suggests the selected features are not substantially affected by the type of classifier that is selected. Thus, a good ranker model performs well no matter which classifier we use afterwards.
\end{enumerate}

Thus, it seems clear that it is more important to have a good classifier when selecting the features, rather than trying to use the same model for either ranking and classifying. 

\subsubsection{Effect of overfitting in the FS step}

\begin{figure}
	\centering
	\begin{minipage}[]{0.95\linewidth}
	    \centering
 		\includegraphics[width=0.95\textwidth]{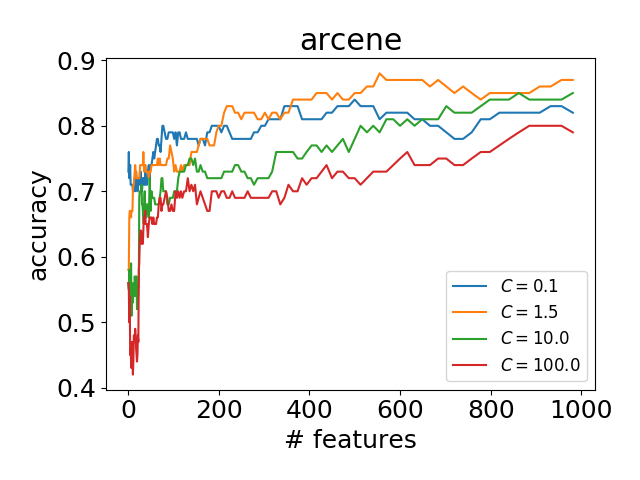}\\
	\end{minipage}
\caption{
 Effect of the overfitting problem during the FS step. As the classifier increases its overfitting, the quality of the selection decreases substantially. 
}
\label{fig:overfitting}
\end{figure}

As we have seen in the previous subsection, a good classifier selection is crucial in order to obtain a solid feature selection, and thus we also would like to evaluate how the overfitting of a model might affect the selection of features. In this case, we have only used the Arcene dataset to illustrate the scenario, as it provides very meaningful results. 

In the previous subsection, it can be seen that the best results for Arcene dataset were achieved using SVM-RBF for both ranking and classifying, with $C = 1.5$ as meta-parameter. Thus, our experiment consists in, using the same classifier configuration, check how the result varies as we modify the $C$ parameter during the FS step. Our algorithm meta-parameters were fixed to their minimum values ($\gamma = 0$ and $reps = 1$).

Fig. \ref{fig:overfitting} shows the obtained results. The best results are achieved when using the same $C$ value that was employed also during the classification step. It can be seen that using a lower value does not substantially affect the result. However, when introducing overfitting (high $C$ values), the quality of the result is drastically reduced.

Thus, we strongly recommend, during the FS step, the use of classifiers that are able to prevent overfitting, as this is a problem that could introduce noise in the FS ranking system.





\section{A comparative study of our proposal with other FS methods}
\label{sec:results}

In order to test the adequacy of our proposal, we have compared it against the most representative feature selection methods available nowadays, namely: (a) both Sklearn's LASSO and Elastic Net implementations \cite{kim2007interior,friedman2010regularization}; (b) the MIM revision developed in \cite{ross2014mutual}, also implemented in the Sklearn package; (c) the ReliefF algorithm \cite{kononenko1994estimating} implemented in the Skrebate package \cite{urbanowicz2017benchmarking}; and (d) the DFS algorithm \cite{li2016deep}, using our own Keras implementation. In this last case, and in order to make a fair comparison, we have used the DFS mask values as the feature ranking. This approach was not mentioned in the original paper mentioned above \cite{li2016deep}, as they just train the model with the new mask and constraint (like LASSO and Elastic Net behavior). However, experimental results show that this approach achieves better results when using our variation. The code with our implementations is available in GitHub\footnote{https://github.com/braisCB/SFS}.

\subsection{NIPS 2003 Feature Selection Challenge}

As the number of instances is relatively low in all datasets, we have followed the same approach explained in the ablation study described in the previous section, that is, we have used four SVM variants to test the performance of every algorithm. In the case of the DFS algorithm, we have used the three-layer NN that was also explained in the previous section as the feature selection algorithm. We also have tried the same network with our proposed approach.

\subsubsection{Arcene Dataset}
\begin{table*}
	\centering
	\caption{Arcene accuracy results, and in parenthesis, number of features used to achieve the best score. The baseline is the classifier accuracy without removing any feature. Same ranker means the same SVM model is used both as ranker and as classifier. Bold shows the result of the best ranker-classifier combination. }
	\label{tab:arcene}
	\begin{adjustbox}{max width=\columnwidth}
	\begin{tabular}{| c | c | c | c | c | c |} 
      	\hline
      	\multirow{2}{*}{FS Method} & \multirow{2}{*}{Ranker} & \multicolumn{4}{|c|}{SVM Classifier (\# of features)}\\
      	\cline{3-6} 
      	  & & Linear & Poly & RBF & Sigmoid \\
      	\hline
      	Baseline (all features) & --- & 83.0 & 87.0 & 72.0 & 69.0 \\
      	\hline
      	LASSO \cite{kim2007interior,friedman2010regularization} & --- & 70.0 & --- & --- & --- \\
      	\hline
      	Elastic Net \cite{kim2007interior,friedman2010regularization} & --- & 76.0 & --- & --- & --- \\
      	\hline
      	MIM \cite{ross2014mutual} & --- & 83.0 (84) & \textbf{88.0} (8164) & 81.0 (25) & 69.0 (1817) \\
      	\hline
      	ReliefF \cite{kononenko1994estimating} & --- & 83.0 (10000) & \textbf{88.0} (8818) & 75.0 (600) & 71.0 (2231) \\
      	\hline
      	DFS \cite{zou2015deep} & NN & 86.0 (1145) & 87.0 (8810) & 77.0 (302) & 72.0 (4439) \\
      	\hline
      	\hline
        \multirow{5}{*}{SFS} & Same ($\gamma = 0$) & 84.0 (4328) & 87.0 (10000) & 83.0 (600) & 74.0 (3715) \\
      	\cline{2-6} 
         & Best ($\gamma = 0$) & 85.0 (318) & 87.0 (9036) & 86.0 (513) & 78.0 (2601) \\
      	\cline{2-6} 
      	 & Same ($\gamma = 0.975$) & 83.0 (3189) & \textbf{88.0} (9760) & \textbf{88.0} (487) & 72.0 (3715) \\
      	\cline{2-6} 
         & Best ($\gamma = 0.975$) & 84.0 (1145) & \textbf{88.0} (9750) & \textbf{88.0} (487) & 84.0 (1145) \\
      	\cline{2-6} 
         & NN ($\gamma = 0.9$, reps $ = 3$) & 83.0 (1599) & 87.0 (5869) & \textbf{88.0 (19)} & 75.0 (465) \\
      	\hline
    \end{tabular}
    \end{adjustbox}
\end{table*}

Table \ref{tab:arcene} shows the obtained results. Overall, the best results are achieved when using the polynomial kernel. However, for all the other tested algorithms but our approach, these are achieved when using practically almost all features (in the most favourable case, 81,6\% of the total features). Differently, our algorithm is able to achieve the best same results in accuracy as the polynomial kernel, but  when using a RBF kernel and only $19$ features, that is, the 4,87\% of the whole features of the dataset.

\subsubsection{Dexter Dataset}

\begin{table*}
	\centering
	\caption{Dexter accuracy results, and in parenthesis, number of features used to achieve the best score. The baseline is the classifier accuracy without removing any feature. Same ranker means the same SVM model is used both as ranker and as classifier. Best shows the result of the best ranker-classifier combination.}
	\label{tab:dexter}
	\begin{adjustbox}{max width=\columnwidth}
	\begin{tabular}{| c | c | c | c | c | c |} 
      	\hline
      	  \multirow{2}{*}{FS Method} & \multirow{2}{*}{Ranker} & \multicolumn{4}{|c|}{SVM Classifier (\# of features)}\\
      	\cline{3-6} 
      	  & & Linear & Poly & RBF & Sigmoid \\
      	\hline
      	Baseline (all features) & --- & 93.7 & 89.0 & 89.0 & 89.0 \\
      	\hline
      	LASSO \cite{kim2007interior,friedman2010regularization} & --- & 89.0 & --- & --- & --- \\
      	\hline
      	Elastic Net \cite{kim2007interior,friedman2010regularization} & --- & 91.3 & --- & --- & --- \\
      	\hline
      	MIM \cite{ross2014mutual} & --- & 93.7 (20000) & 92.0 (166) & 91.0 (152) & 90.7 (1533) \\
      	\hline
      	ReliefF \cite{kononenko1994estimating} & --- & 93.7 (20000) & 91.7 (31) & 91.0 (68) & 90.0 (111) \\
      	\hline
      	DFS \cite{zou2015deep} & NN & 93.7 (20000) & 90.7 (93) & 90.3 (2979) & 90.3 (3746) \\
      	\hline
      	\hline
        \multirow{5}{*}{SFS} & Same ($\gamma = 0$) & 94.0 (939) & 92.0 (56) & 91.3 (52) & 91.0 (4710) \\
      	\cline{2-6} 
         & Best ($\gamma = 0$) & 94.0 (939) & 92.7 (52) & 91.3 (52) & 91.0 (4710) \\
      	\cline{2-6} 
      	 & Same ($\gamma = 0.975$) & \textbf{94.3 (1419)} & 91.0 (66) & 92.7 (37) & 92.0 (64) \\
      	\cline{2-6} 
         & Best ($\gamma = 0.975$) & \textbf{94.3 (1419)} & 93.7 (7071) & 94.0 (7071) & 93.7 (3384) \\
      	\cline{2-6} 
         & NN ($\gamma = 0.9$, reps $ = 3$) & 94.0 (2493) & 91.7 (111) & 91.7 (114) & 91.0 (124) \\
      	\hline
    \end{tabular}
    \end{adjustbox}
\end{table*}

The problem found with the polynomial kernel in the Arcene dataset appears also in Dexter dataset, although this time with the Linear kernel, as we can see in Table \ref{tab:dexter}. Again, our algorithm outperforms the other techniques, independently of the SVM kernel selected. This time our proposal not only reduces considerably the number of features used (100\% for the best result in other methods, while ours use 7,1\%), but also improves slightly the accuracy result obtained.

\subsubsection{Dorothea Dataset}

\begin{table*}
	\centering
	\caption{Dorothea accuracy results, in parenthesis, number of features used to achieve the best score. The baseline is the classifier accuracy without removing any feature. Same ranker means the same SVM model is used both as ranker and as classifier. Best shows the result of the best ranker-classifier combination.}
	\label{tab:dorothea}
	\begin{adjustbox}{max width=\columnwidth}
	\begin{tabular}{| c | c | c | c | c | c |} 
      	\hline
      	  \multirow{2}{*}{FS Method} & \multirow{2}{*}{Ranker} & \multicolumn{4}{|c|}{SVM Classifier (\# of features)}\\
      	\cline{3-6} 
      	  & & Linear & Poly & RBF & Sigmoid \\
      	\hline
      	Baseline (all features) & --- & 93.1 & 90.3 & 9.7 & 92.3 \\
      	\hline
      	LASSO \cite{kim2007interior,friedman2010regularization} & --- & 93.4 & --- & --- & --- \\
      	\hline
      	Elastic Net \cite{kim2007interior,friedman2010regularization} & --- & 93.7 & --- & --- & --- \\
      	\hline
      	MIM \cite{ross2014mutual} & --- & 94.0 (1677) & 94.9 (112) & 94.9 (77) & 94.9 (77) \\
      	\hline
      	ReliefF \cite{kononenko1994estimating} & --- & 94.6 (1030) & 94.3 (79) & 94.3 (67) & 94.3 (7) \\
      	\hline
      	DFS \cite{zou2015deep} & NN & 94.6 (59) & 95.1 (61) & \textbf{95.4 (57)} & 95.1 (53) \\
      	\hline
      	\hline
        \multirow{5}{*}{SFS} & Same ($\gamma = 0$) & 94.0 (3995) & \textbf{95.4} (283) & 95.1 (138) & 94.9 (94) \\
      	\cline{2-6} 
         & Best ($\gamma = 0$) & 94.3 (94) & \textbf{95.4} (283) & \textbf{95.4} (381) & 94.9 (94) \\
      	\cline{2-6} 
      	 & Same ($\gamma = 0.975$) & 94.9 (222) & 95.1 (88) & 94.6 (23) & 94.9 (27) \\
      	\cline{2-6} 
         & Best ($\gamma = 0.975$) & 94.9 (69) & 95.1 (88) & 94.9 (65) & 95.1 (11318) \\
      	\cline{2-6} 
         & NN ($\gamma = 0.9$, reps $ = 3$) & 94.3 (193) & 94.3 (85) & 94.9 (381) & 94.6 (402) \\
      	\hline
    \end{tabular}
    \end{adjustbox}
\end{table*}

Table \ref{tab:dorothea} shows that the best result is achieved when using the DFS algorithm with an RBF kernel. However, our approach is able to achieve the same accuracy, both with Polynomial and RBF kernels, but requiring  more features (57 for the DFS algorithm and 283 for our proposal, respectively 0,057\% and 0,3\% regarding the complete set of features).

\subsubsection{Gisette Dataset}

\begin{table*}
	\centering
	\caption{Gisette accuracy results, in parenthesis number of features used to achieve the best score. The baseline is the classifier accuracy without removing any feature. Same ranker means the same SVM model is used both as ranker and as classifier. Best shows the result of the best ranker-classifier combination.}
	\label{tab:gisette}
	\begin{adjustbox}{max width=\columnwidth}
	\begin{tabular}{| c | c | c | c | c | c |} 
      	\hline
      	  \multirow{2}{*}{FS Method} & \multirow{2}{*}{Ranker} & \multicolumn{4}{|c|}{SVM Classifier (\# of features)}\\
      	\cline{3-6} 
      	  & & Linear & Poly & RBF & Sigmoid \\
      	\hline
      	Baseline (all features) & --- & 97.7 & 97.5 & 96.9 & 95.7 \\
      	\hline
      	LASSO \cite{kim2007interior,friedman2010regularization} & --- & 97.4 & --- & --- & --- \\
      	\hline
      	Elastic Net \cite{kim2007interior,friedman2010regularization} & --- & 97.4 & --- & --- & --- \\
      	\hline
      	MIM \cite{ross2014mutual} & --- & 97.7 (2212) & 97.8 (1997) & 97.6 (902) & 96.7 (645) \\
      	\hline
      	ReliefF \cite{kononenko1994estimating} & --- & 98.2 (3083) & 98.2 (1803) & 97.7 (1850) & 97.1 (1587) \\
      	\hline
      	DFS \cite{zou2015deep} & NN & 98.2 (168) & 98.1 (3503) & \textbf{98.3} (130) & 97.5 (163) \\
      	\hline
      	\hline
        \multirow{5}{*}{SFS} & Same ($\gamma = 0$) & 98.1 (551) & 98.1 (333) & 98.0 (423) & 97.8 (645) \\
      	\cline{2-6} 
         & Best ($\gamma = 0$) & 98.1 (299) & 98.3 (662) & 98.1 (275) & 97.8 (412) \\
      	\cline{2-6} 
      	 & Same ($\gamma = 0.975$) & 98.3 (1080) & 98.2 (360) & 98.2 (412) & 98.0 (351) \\
      	\cline{2-6} 
         & Best ($\gamma = 0.975$) & 98.3 (1080) & 98.3 (483) & 98.3 (351) & 98.0 (351) \\
      	\cline{2-6} 
         & NN ($\gamma = 0.9$, reps $ = 3$) & 98.0 (1397) & \textbf{98.4 (291)} & 98.2 (333) & 97.8 (324) \\
      	\hline
    \end{tabular}
    \end{adjustbox}
\end{table*}

Table \ref{tab:gisette} shows that our algorithm achieves the highest score when using the Neural Network as ranker, although DFS obtains similar accuracy with fewer features (130 for DFS versus 291 for our approach, respectively 2,6\% and 5,8\% of the total set of features). Compared with either MIM or ReliefF, our algorithm is able to systematically select fewer features.

\subsubsection{Madelon Dataset}

\begin{table*}
	\centering
	\caption{Madelon accuracy results. In parenthesis, number of features used to achieve the best score. The baseline is the classifier accuracy without removing any feature. Same ranker means the same SVM model is used both as ranker and as classifier. Best shows the result of the best ranker-classifier combination.}
	\label{tab:madelon}
	\begin{adjustbox}{max width=\columnwidth}
	\begin{tabular}{| c | c | c | c | c | c |} 
      	\hline
      	  \multirow{2}{*}{FS Method} & \multirow{2}{*}{Ranker} & \multicolumn{4}{|c|}{SVM Classifier (\# of features)}\\
      	\cline{3-6} 
      	  & & Linear & Poly & RBF & Sigmoid \\
      	\hline
      	Baseline (all features) & --- & 53.0 & 67.7 & 68.7 & 52.3 \\
      	\hline
      	LASSO \cite{kim2007interior,friedman2010regularization} & --- & 58.3 & --- & --- & --- \\
      	\hline
      	Elastic Net \cite{kim2007interior,friedman2010regularization} & --- & 59.7 & --- & --- & --- \\
      	\hline
      	MIM \cite{ross2014mutual} & --- & 62.5 (5) & 72.5 (128) & 80.2 (15) & 57.2 (3) \\
      	\hline
      	ReliefF \cite{kononenko1994estimating} & --- & 62.7 (4) & 74.7 (36) & \textbf{91.5 (9)} & 53.2 (111) \\
      	\hline
      	DFS \cite{zou2015deep} & NN & 62.5 (8) & 72.0 (40) & 90.8 (11) & 53.0 (1) \\
      	\hline
      	\hline
        \multirow{5}{*}{SFS} & Same ($\gamma = 0$) & 59.2 (56) & 73.5 (42) & 83.9 (32) & 54.7 (76) \\
      	\cline{2-6} 
         & Best ($\gamma = 0$) & 62.3 (1) & 73.5 (42) & 83.9 (32) & 62.3 (58) \\
      	\cline{2-6} 
      	 & Same ($\gamma = 0.975$) & 58.2 (218) & 76.2 (18) & 89.3 (12) & 52.3 (500) \\
      	\cline{2-6} 
         & Best ($\gamma = 0.975$) & 62.0 (4) & 76.2 (18) & 90.8 (17) & 57.0 (374) \\
      	\cline{2-6} 
         & NN ($\gamma = 0.9$, reps $ = 3$) & 62.8 (9) & 77.8 (16) & \textbf{91.5} (25) & 52.8 (87) \\
      	\hline
    \end{tabular}
    \end{adjustbox}
\end{table*}

The ReliefF algorithm achieves the best score, as Table \ref{tab:madelon} shows. However, we may note that our algorithm is also able to achieve the same score, but using more features (1,8\% and 5\% of the total number of features).

\vspace{0.2cm}

To sum up, our algorithm is able to achieve the same, or even improve slightly, the accuracy results compared with those obtained by the state-of-the art algorithms for all NIPS datasets. Regarding the number of features, our proposal behaves extraordinarily in some datasets, in which the reduction of the features needed is remarkable, while in other cases in which the existing methods accomplish an important reduction our approach needs approximately the double. 

\subsection{Regression}

We conducted two experiments to show the behavior of our SFS algorithm in a regression problem. As early mentioned, we have used both the \textit{Relative location of CT slices on axial axis} and the \textit{Energy Molecule} dataset . To perform the experiments, we selected the same approach presented in Section \ref{sec:reps}: a 3-layer NN ($150$, $100$ and $50$ nodes, respectively), along with BN and ReLU activation function. Both differ only in the output (now it is just one node) and the loss function (MSE). A 5-fold cross-validation was performed.

\begin{table}
	\centering
	\caption{SFS performance in Regression problems. Relative location of CT slices on axial axis Dataset MAE results.}
	\label{tab:slice_localization_data}
	\begin{adjustbox}{max width=\columnwidth}
	\begin{tabular}{| c | c | c | c | c |} 
      	\hline
      	  \multirow{2}{*}{FS Method} & \multicolumn{4}{|c|}{\# of features}\\
      	\cline{2-5} 
      	  & 19 & 38 & 96 & 192 \\
      	\hline
      	DFS \cite{zou2015deep}  & 4.18 & \textbf{2.84} & 2.32 & 2.39 \\
      	\hline
      	SFS ($\gamma = 0$) & 6.55 & 3.70 & 2.71 & \textbf{2.24} \\
      	\hline
      	SFS ($\gamma = 0.9$) & 4.18 & 3.16 & 2.60 & 2.31 \\
      	\hline
      	SFS$+$DFS ($\gamma = 0$) & 4.41 & 2.94 & 2.25 & 2.40 \\
      	\hline
      	SFS$+$DFS ($\gamma = 0.9$) & \textbf{4.08} & 2.87 & \textbf{2.30} & 2.27 \\
      	\hline
    \end{tabular}
    \end{adjustbox}
\end{table}

Table \ref{tab:slice_localization_data} shows the obtained results for the \textit{Relative location of CT slices on axial axis} dataset. We compare our approach with the DFS algorithm, using an input mask with $l_1 = 5 \cdot 10^{-4}$ as weight penalty. Additionally, we also consider using our SFS method adding the DFS mask at the input. We refer to that configuration as SFS$+$DFS. We have fixed the parameter \emph{reps} to $2$ in all experiments. The results show that SFS and the combination SFS$+$DFS, both with a 3-layer CNN, achieved the best results. For the SFS alone, 192 features, that is the whole set of features were needed, while for the combination of DFS with our approach, only 96 (50\%) were needed for an almost equal accuracy.

\begin{table}
	\centering
	\caption{SFS performance in Regression problems. Energy Molecule Dataset MAE results.}
	\label{tab:energy_molecule}
	\begin{adjustbox}{max width=\columnwidth}
	\begin{tabular}{| c | c | c | c | c |} 
      	\hline
      	  \multirow{2}{*}{FS Method} & \multicolumn{4}{|c|}{\# of features}\\
      	\cline{2-5} 
      	  & 63 & 127 & 318 & 637 \\
      	\hline
      	DFS \cite{zou2015deep}  & \textbf{0.139} & 0.136 & 0.137 & 0.142 \\
      	\hline
      	SFS ($\gamma = 0$) & 0.292 & 0.215 & 0.146 & 0.141 \\
      	\hline
      	SFS ($\gamma = 0.9$) & 0.145 & \textbf{0.132} & \textbf{0.131} & \textbf{0.136} \\
      	\hline
      	SFS$+$DFS ($\gamma = 0$) & 0.145 & 0.139 & 0.143 & 0.149 \\
      	\hline
      	SFS$+$DFS ($\gamma = 0.9$) & \textbf{0.139} & 0.136 & \textbf{0.132} & \textbf{0.138} \\
      	\hline
    \end{tabular}
    \end{adjustbox}
\end{table}

Using the same configuration, Table \ref{tab:energy_molecule} shows the obtained results for the \textit{Energy Molecule} dataset. The results show that SFS with $\gamma = 0.9$ is able to achieve almost the best score by just using 127 features (the 10\% of the whole dataset). Either SFS or SFS$+$DFS achieve the best scores in all different configurations.

\subsection{Results of the approach in Big Data scenarios}

One of the main advantages of our method is that it is possible to use it in Big Data environments, as it was developed to be used in state-of-the-art architectures like Convolutional Neural Networks. To that end, we have tested the behavior of our approach over four different datasets using the Wide Residual Network WRN-16-4 \cite{zagoruyko2016wide} as classifier.

\begin{table}
	\centering
	\caption{MNIST accuracy results using a WRN-16-4 \cite{zagoruyko2016wide} as classifiers.}
	\label{tab:mnist}
	\begin{adjustbox}{max width=\columnwidth}
	\begin{tabular}{| c | c | c | c | c | c |} 
      	\hline
      	  \multirow{2}{*}{FS Method} & \multirow{2}{*}{Ranker} & \multicolumn{4}{|c|}{\# of features}\\
      	\cline{3-6} 
      	  & & 39 & 78 & 196 & 392 \\
      	\hline
      	DFS \cite{zou2015deep}  & 3-layer CNN & 95.73 & 98.56 & 99.32 & 99.46 \\
      	\hline
      	DFS \cite{zou2015deep}  & WRN-16-4 & 92.92 & 97.36 & 98.72 & 98.98 \\
      	\hline
      	SFS & 3-layer CNN ($\gamma = 0$) & 89.78 & 95.66 & 99.14 & 99.53 \\
      	\hline
      	SFS & 3-layer CNN ($\gamma = 0.9$) & \textbf{97.08} & 98.49 & 99.13 & \textbf{99.56} \\
      	\hline
      	SFS & WRN-16-4 ($\gamma = 0$) & 88.18 & 94.98 & 98.70 & 99.41 \\
      	\hline
      	SFS & WRN-16-4 ($\gamma = 0.9$) & 96.76 & \textbf{98.62} & 99.10 & 99.30 \\
      	\hline
      	SFS$+$DFS & 3-layer CNN ($\gamma = 0$) & 95.60 & 98.47 & \textbf{99.38} & 99.48 \\
      	\hline
      	SFS$+$DFS & WRN-16-4 ($\gamma = 0$) & 92.92 & 97.36 & 98.72 & 98.98 \\
      	\hline
    \end{tabular}
    \end{adjustbox}
\end{table}

As ranking method, we have tried two different configurations: the previously mentioned WRN-16-4 and a standard 3-layer CNN. It consists in two convolutional layers with 16 and 32 channels, respectively. After each convolution, a $2\times2$ Max-Pooling Layer. Finally, two Fully Connected layers are used: the first one contains 1024 nodes, and the last one has the size of the number of labels in the dataset ($100$ for CIFAR-100, $10$ for the others). Both Batch Normalization (DN) \cite{ioffe2015batch} and ReLU activation function are applied right after all hidden layers, and the Softmax function is applied to the output. The Adam optimizer \cite{kingma2014adam} is used to train the model.

\begin{table}
	\centering
	\caption{Fashion-MNIST accuracy results using a WRN-16-4 \cite{zagoruyko2016wide} as classifiers.}
	\label{tab:fashion_mnist}
	\begin{adjustbox}{max width=\columnwidth}
	\begin{tabular}{| c | c | c | c | c | c |} 
      	\hline
      	  \multirow{2}{*}{FS Method} & \multirow{2}{*}{Ranker} & \multicolumn{4}{|c|}{\# of features}\\
      	\cline{3-6} 
      	  & & 39 & 78 & 196 & 392 \\
      	\hline
      	DFS \cite{zou2015deep}  & 3-layer CNN & 78.85 & 85.5 & \textbf{90.45} & 92.41 \\
      	\hline
      	DFS \cite{zou2015deep}  & WRN-16-4 & 72.58 & 76.52 & 87.05 & \textbf{92.61} \\
      	\hline
      	SFS & 3-layer CNN ($\gamma = 0$) & 67.85 & 81.86 & 89.33 & 92.36 \\
      	\hline
      	SFS & 3-layer CNN ($\gamma = 0.9$) & \textbf{82.63} & \textbf{86.33} & 90.09 & \textbf{92.60} \\
      	\hline
      	SFS & WRN-16-4 ($\gamma = 0$) & 64.17 & 73.53 & 85.60 & 92.48 \\
      	\hline
      	SFS & WRN-16-4 ($\gamma = 0.9$) & 77.99 & 82.58 & 88.16 & 91.72 \\
      	\hline
      	SFS$+$DFS & 3-layer CNN ($\gamma = 0$) & 79.81 & \textbf{86.29} & \textbf{90.44} & \textbf{92.59} \\
      	\hline
      	SFS$+$DFS & WRN-16-4 ($\gamma = 0$) & 65.09 & 73.87 & 85.86 & 92.40 \\
      	\hline
    \end{tabular}
    \end{adjustbox}
\end{table}

In both networks we use the categorical cross-entropy as loss function, together with an $l_2 = 5 \cdot 10^{-4}$ weight penalty. 

\begin{table}
	\centering
	\caption{CIFAR-10 accuracy results using a WRN-16-4 \cite{zagoruyko2016wide} as classifiers.}
	\label{tab:cifar10}
	\begin{adjustbox}{max width=\columnwidth}
	\begin{tabular}{| c | c | c | c | c | c |} 
      	\hline
      	  \multirow{2}{*}{FS Method} & \multirow{2}{*}{Ranker} & \multicolumn{4}{|c|}{\# of features}\\
      	\cline{3-6} 
      	  & & 153 & 307 & 768 & 1536 \\
      	\hline
      	DFS \cite{zou2015deep}  & 3-layer CNN & 67.43 & \textbf{79.92} & 87.71 & 90.69 \\
      	\hline
      	DFS \cite{zou2015deep}  & WRN-16-4 & 61.51 & 71.19 & 83.94 & 89.47 \\
      	\hline
      	SFS & 3-layer CNN ($\gamma = 0$) & 61.00 & 72.49 & 84.31 & 89.31 \\
      	\hline
      	SFS & 3-layer CNN ($\gamma = 0.9$) & 63.52 & 77.96 & 89.60 & 91.5 \\
      	\hline
      	SFS & WRN-16-4 ($\gamma = 0$) & 53.03 & 60.42 & 85.55 & 90.44 \\
      	\hline
      	SFS & WRN-16-4 ($\gamma = 0.9$) & 64.15 & 79.27 & \textbf{89.85} & \textbf{91.58} \\
      	\hline
      	SFS$+$DFS & 3-layer CNN ($\gamma = 0$) & \textbf{68.13} & 79.03 & 88.04 & 91.06 \\
      	\hline
      	SFS$+$DFS & WRN-16-4 ($\gamma = 0$) & 54.05 & 68.65 & 82.47 & 89.54 \\
      	\hline
    \end{tabular}
    \end{adjustbox}
\end{table}

We have conducted experiments in four well-known image databases described in sections above: MNIST, Fashion-MNIST, CIFAR-10 and CIFAR-100.

\begin{table}
	\centering
	\caption{CIFAR-100 accuracy results using a WRN-16-4 \cite{zagoruyko2016wide} as classifiers.}
	\label{tab:cifar100}
	\begin{adjustbox}{max width=\columnwidth}
	\begin{tabular}{| c | c | c | c | c | c |} 
      	\hline
      	  \multirow{2}{*}{FS Method} & \multirow{2}{*}{Ranker} & \multicolumn{4}{|c|}{\# of features}\\
      	\cline{3-6} 
      	  & & 153 & 307 & 768 & 1536 \\
      	\hline
      	DFS \cite{zou2015deep} & 3-layer CNN & 34.55 & \textbf{49.68} & 57.92 & 62.22 \\
      	\hline
      	DFS \cite{zou2015deep} & WRN-16-4 & 28.24 & 40.77 & 56.98 & \textbf{67.42} \\
      	\hline
      	SFS & 3-layer CNN ($\gamma = 0$) & 24.53 & 34.14 & 52.67 & 63.45 \\
      	\hline
      	SFS & 3-layer CNN ($\gamma = 0.9$) & 30.74 & 44.55 & 61.49 & 66.96 \\
      	\hline
      	SFS & WRN-16-4 ($\gamma = 0$) & 24.66 & 37.86 & 56.66 & 66.39 \\
      	\hline
      	SFS & WRN-16-4 ($\gamma = 0.9$) & 27.88 & 44.45 & \textbf{62.83} & 67.22 \\
      	\hline
      	SFS$+$DFS & 3-layer CNN ($\gamma = 0$) & \textbf{36.86} & 46.64 & 60.46 & 63.55 \\
      	\hline
      	SFS$+$DFS & WRN-16-4 ($\gamma = 0$) & 25.08 & 37.29 & 54.07 & 64.94 \\
      	\hline
    \end{tabular}
    \end{adjustbox}
\end{table}

\subsubsection{MNIST}

No data augmentation techniques were used to train the models. The 3-layer CNN was trained for 40 epochs, while the WRN-16-4 required 80 epochs. Table \ref{tab:mnist} shows the obtained results. As it can be observed, our approach achieved the best scores balancing accuracy and number of features used. It is worth mentioning that the accuracy of the classifier highly improves when using a high $\gamma$ value. In the next section we will show our intuition behind this effect.

\subsubsection{Fashion-MNIST}

In this case, as data augmentation we used random horizontal flips, along with horizontal and vertical random shifts (up to $4$ pixels). As this dataset is more complex than MNIST, we increased the number of training epochs to $80$ in the case of the 3-layer CNN, and $130$ when using the WRN-16-4. Table \ref{tab:fashion_mnist} shows the obtained results. Overall, our SFS approach with $\gamma = 0.9$ achieved the best scores.

\subsubsection{CIFAR-10}

We used for this case the same data augmentation and training configuration that was presented in the previous Fashion-MNIST dataset, but increasing the random shifts up to $5$ pixels. Table \ref{tab:cifar10} shows the obtained results. The use of DFS helps the model to achieve better results whenever the number of kept features is low. On the contrary, a high $\gamma$ value is useful with a higher number of features (more than $25\%$ of the total features).

\subsubsection{CIFAR-100}

We used the same training configuration as in the previous CIFAR-10. Table \ref{tab:cifar100} shows the obtained results. In this dataset our results are not clearly better than DFS. We believe that the overfitting problem is causing this results (although the training accuracy is close to 100\%, the test accuracy never reaches 70\%).

\subsection{Checking Classifier's Reliability}

As we mentioned before, to our knowledge this is the first Feature Selection algorithm, besides Tree-based techniques, that is able to provide the feature importance for each sample. The main advantage of our proposal is that we can provide an explanation about the classifier's decision. However, we would like to focus on a different scenario: checking the classifier's reliability.

If we take a look at the MNIST results (Table \ref{tab:mnist}), we can see that the best results are achieved when using as ranker the 3-layer CNN instead of the WRN-16-4 model. This result was not expected because the latter is a better classifier than the former. Thus, our intuition in this effect was reduced to two points:
\begin{enumerate}
    \item \emph{Max-pooling:} We think the main disadvantage of our algorithm is that it does not manage feature correlation, unless the training algorithm does it. In the case of a CNN model, local correlations can be detected by using Pooling layers. Contrary to the WRN-16-4 classifier, the 3-layer CNN contains 2 Max-pooling layers, which could help our algorithm to achieve better scores.
    \item \emph{Over-fitting:} During the training step, the training set accuracy always reaches a perfect score when using the WRN-16-4. This could led our algorithm to a bad generalization behavior, as explained in the ablation study section (see Fig. \ref{fig:overfitting}). 
\end{enumerate}

Thus, we conclude that the \textit{quality} of the classifier highly affects the behavior of our algorithm. And the MNIST results also show that test accuracy is not a good metric to determine how suitable is a classifier to be used as ranker in our algorithm.

For this reason, we carried on a different experiment, taking advantage of saliency's properties. As we have previously defined, our saliency function is the gradient of the \textit{gain function} with respect to the input. Thus, it measures how we have to modify our input to increase the probability of belonging to the desired class. So, we decided to use our saliency function to do exactly the opposite. The idea is to answer this simple question:
\begin{itemize}
    \item How much do I have to change a sample to change the classifier's output?
\end{itemize}

\begin{table*}
	\centering
	\caption{Adversarial Images. Given an original image $X$, minimal perturbation needed to force a trained classifier to predict each label with more than $95\%$ of certainty.}
	\label{tab:robustness}
	\begin{adjustbox}{max width=\textwidth}
	\setlength{\tabcolsep}{0pt}
	\begin{tabular}{p{2.6cm} c c c c c c c c c c c} 
      	& $X$ & $X = 0$ & $X = 1$ & $X = 2$ & $X = 3$ & $X = 4$ & $X = 5$ & $X = 6$ & $X = 7$ & $X = 8$ & $X = 9$ \\
        \hline
      	WRN-16-4 w/o Input Noise & 
        \begin{minipage}{.11\textwidth}
            \includegraphics[width=\linewidth]{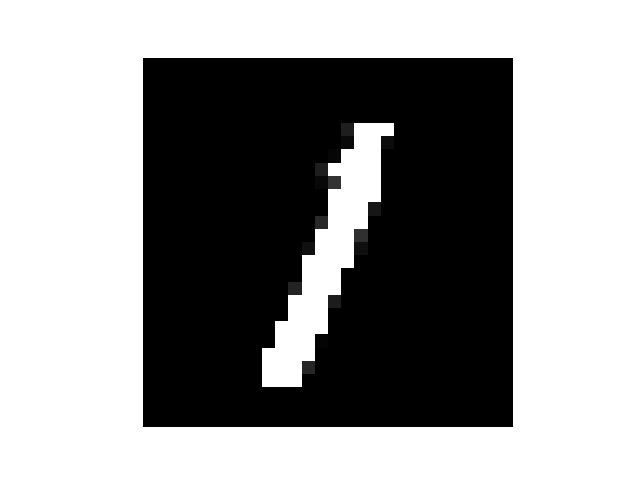}
        \end{minipage} &
        \begin{minipage}{.11\textwidth}
            \includegraphics[width=\linewidth]{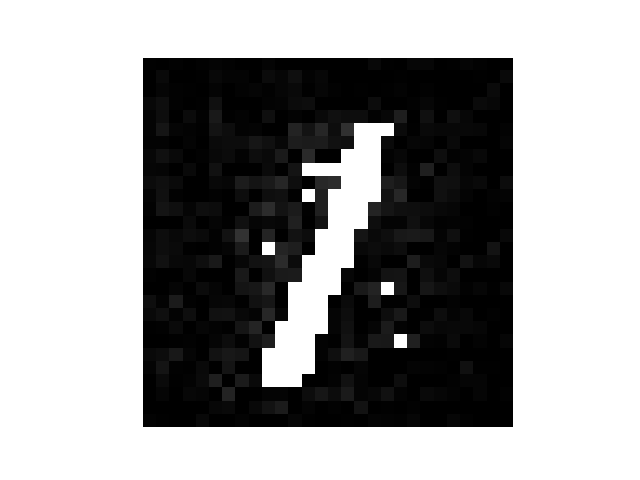}
        \end{minipage} &
        \begin{minipage}{.11\textwidth}
            \includegraphics[width=\linewidth]{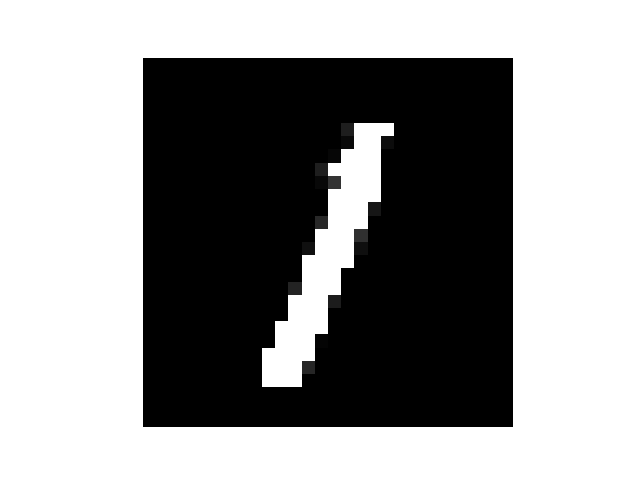}
        \end{minipage} &
        \begin{minipage}{.11\textwidth}
            \includegraphics[width=\linewidth]{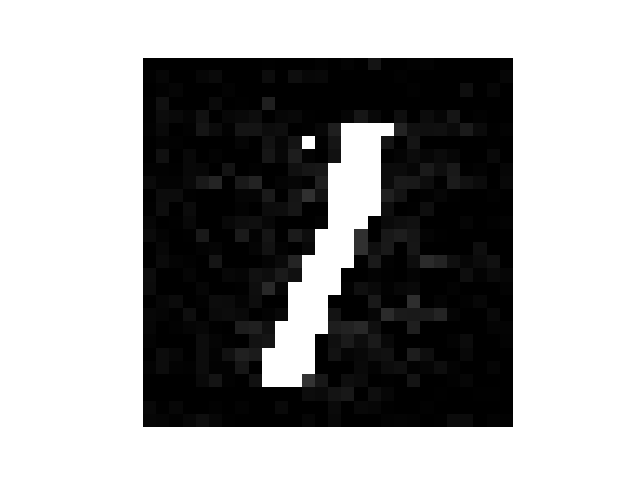}
        \end{minipage} &
        \begin{minipage}{.11\textwidth}
            \includegraphics[width=\linewidth]{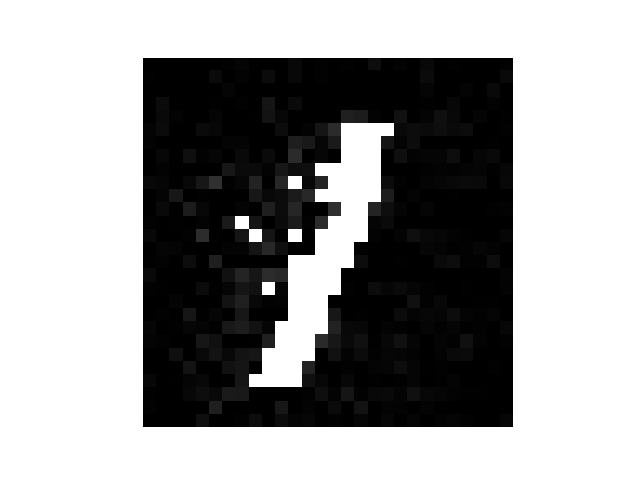}
        \end{minipage} &
        \begin{minipage}{.11\textwidth}
            \includegraphics[width=\linewidth]{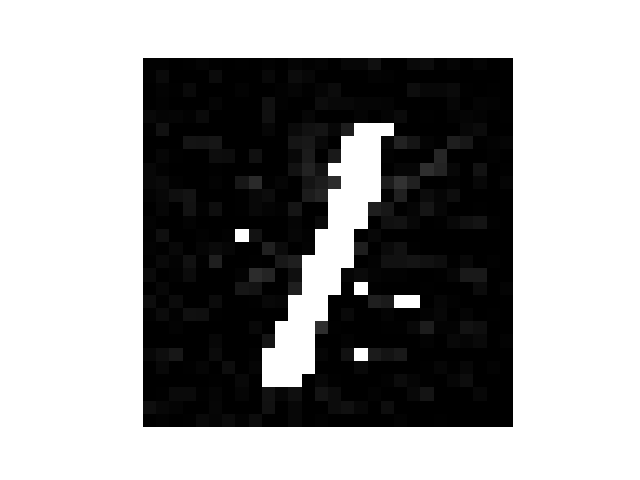}
        \end{minipage} &
        \begin{minipage}{.11\textwidth}
            \includegraphics[width=\linewidth]{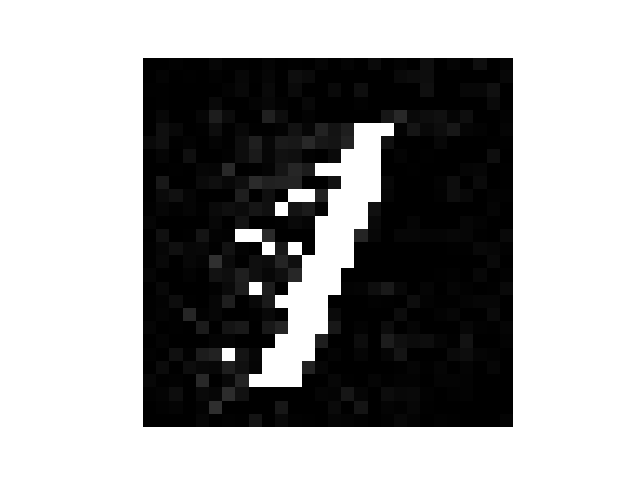}
        \end{minipage} &
        \begin{minipage}{.11\textwidth}
            \includegraphics[width=\linewidth]{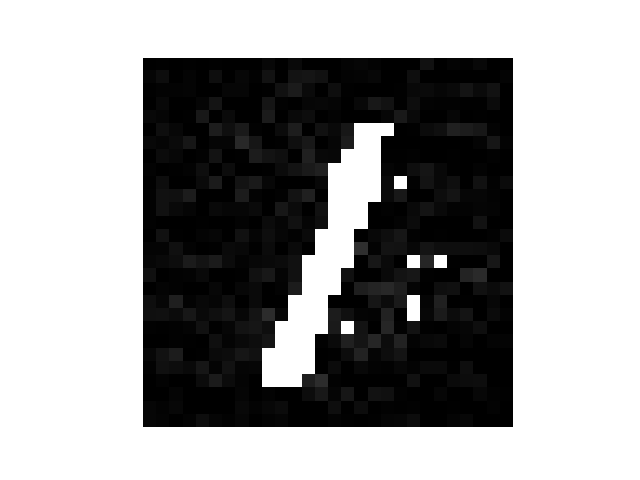}
        \end{minipage} &
        \begin{minipage}{.11\textwidth}
            \includegraphics[width=\linewidth]{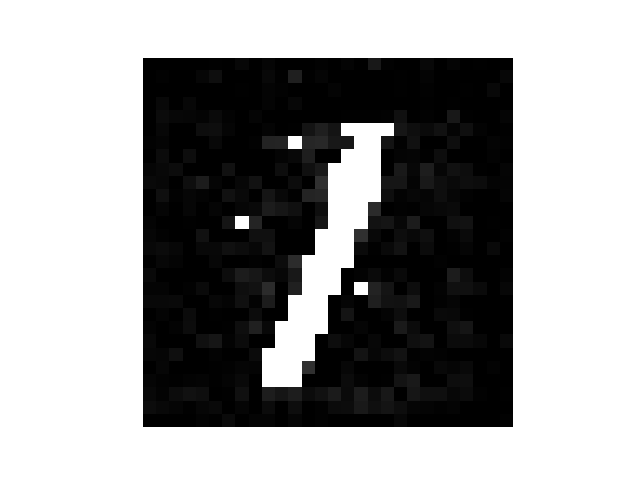}
        \end{minipage} &
        \begin{minipage}{.11\textwidth}
            \includegraphics[width=\linewidth]{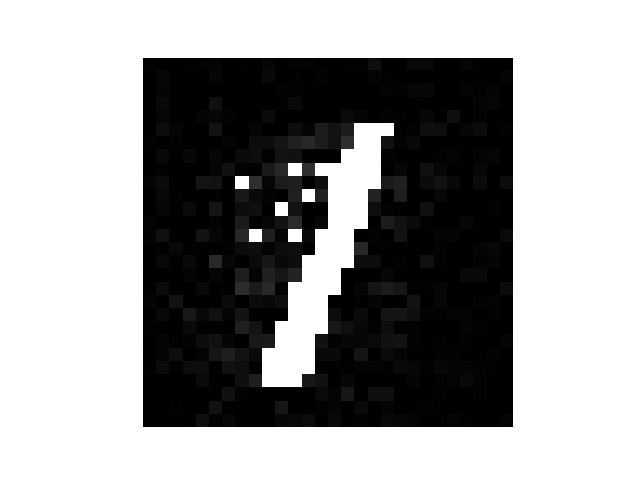}
        \end{minipage} &
        \begin{minipage}{.11\textwidth}
            \includegraphics[width=\linewidth]{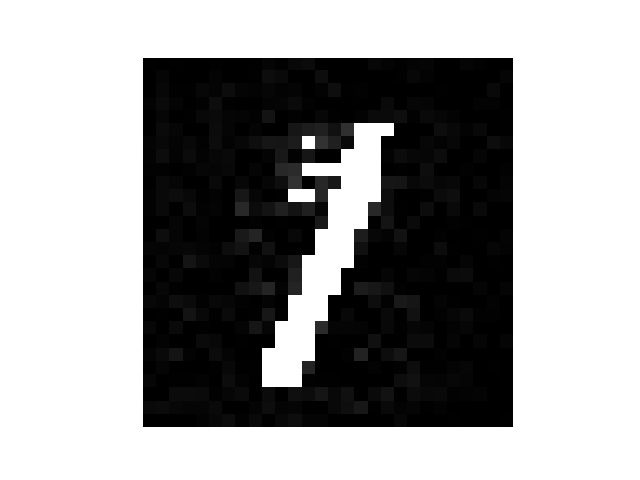}
        \end{minipage} \\
      	WRN-16-4 w/ Input Noise & 
        \begin{minipage}{.11\textwidth}
            \includegraphics[width=\linewidth]{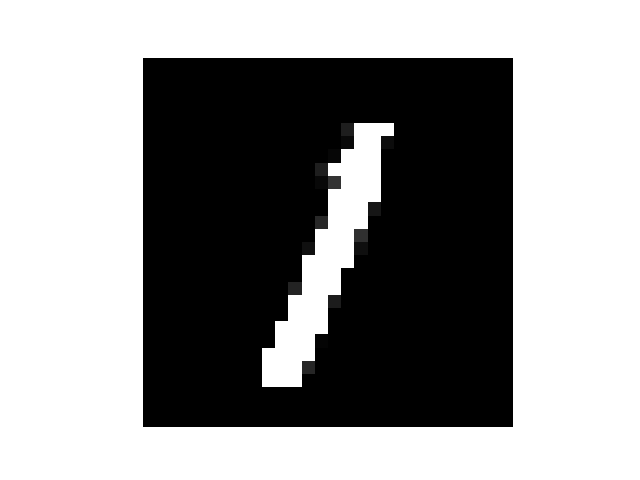}
        \end{minipage} &
        \begin{minipage}{.11\textwidth}
            \includegraphics[width=\linewidth]{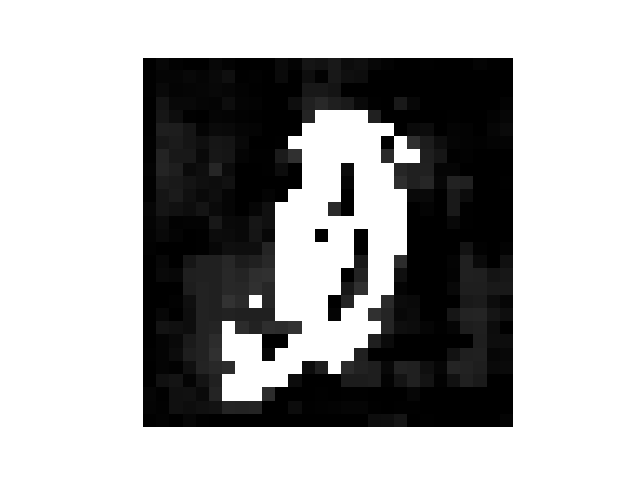}
        \end{minipage} &
        \begin{minipage}{.11\textwidth}
            \includegraphics[width=\linewidth]{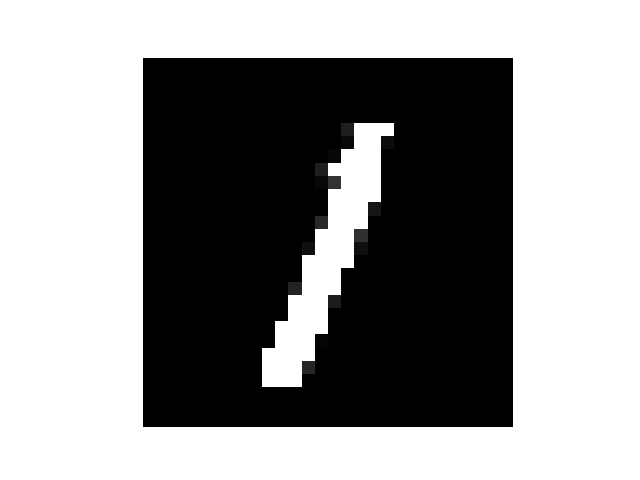}
        \end{minipage} &
        \begin{minipage}{.11\textwidth}
            \includegraphics[width=\linewidth]{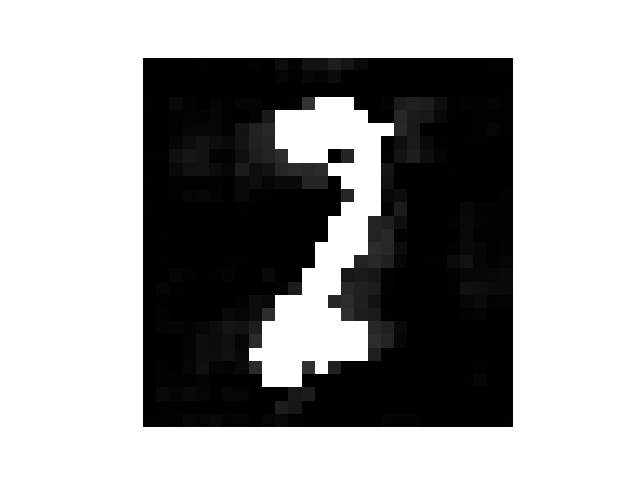}
        \end{minipage} &
        \begin{minipage}{.11\textwidth}
            \includegraphics[width=\linewidth]{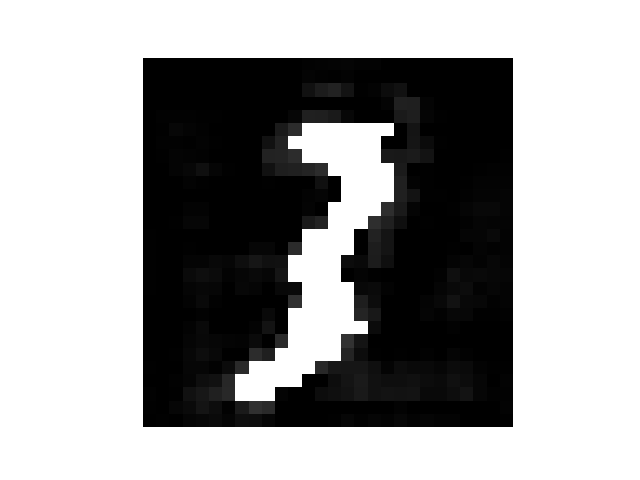}
        \end{minipage} &
        \begin{minipage}{.11\textwidth}
            \includegraphics[width=\linewidth]{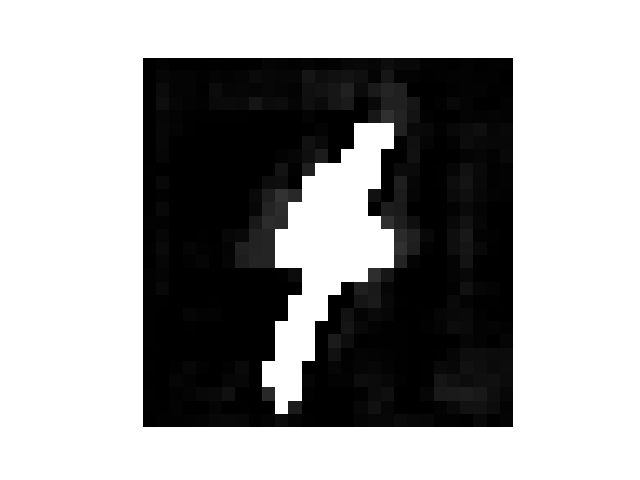}
        \end{minipage} &
        \begin{minipage}{.11\textwidth}
            \includegraphics[width=\linewidth]{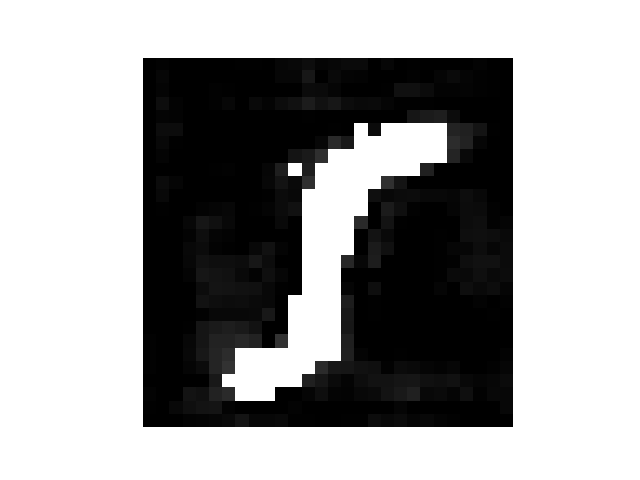}
        \end{minipage} &
        \begin{minipage}{.11\textwidth}
            \includegraphics[width=\linewidth]{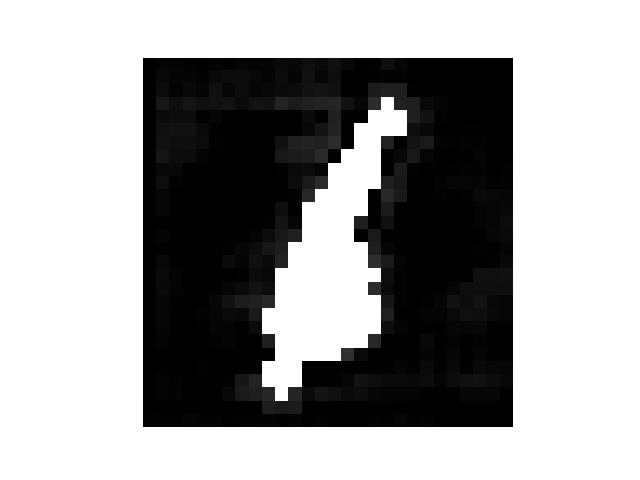}
        \end{minipage} &
        \begin{minipage}{.11\textwidth}
            \includegraphics[width=\linewidth]{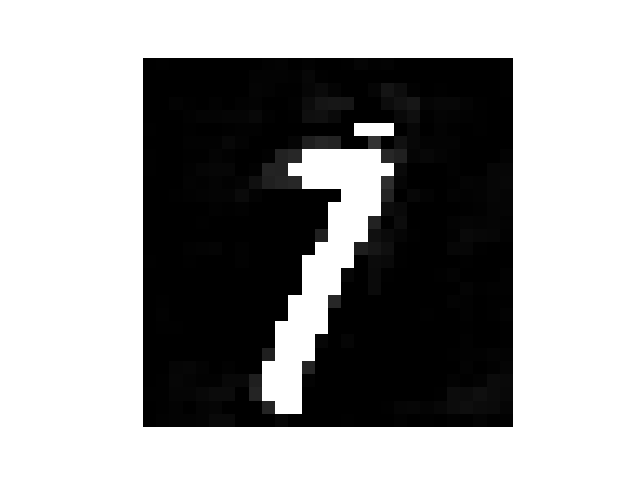}
        \end{minipage} &
        \begin{minipage}{.11\textwidth}
            \includegraphics[width=\linewidth]{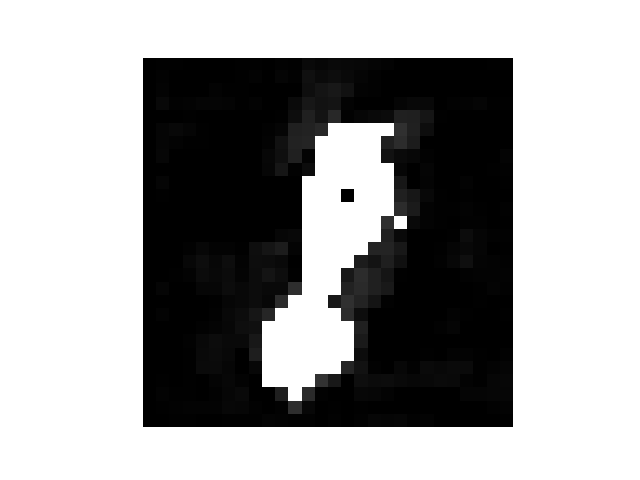}
        \end{minipage} &
        \begin{minipage}{.11\textwidth}
            \includegraphics[width=\linewidth]{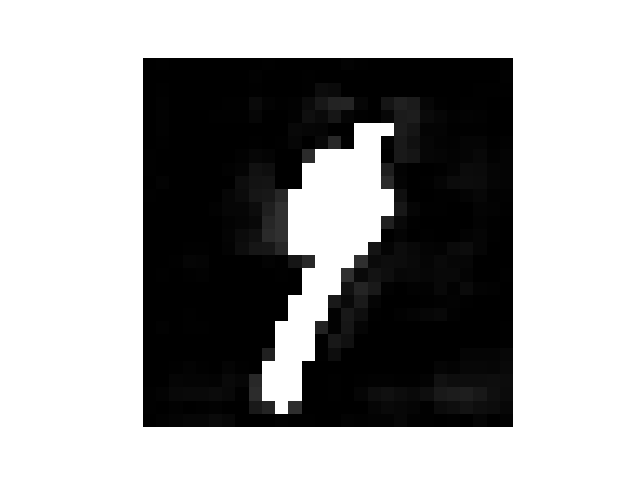}
        \end{minipage} \\
      	
    \end{tabular}
    \end{adjustbox}
\end{table*}

This topic, called \emph{Adversarial Examples} gained a lot of relevance during these years. Several techniques were created in order to increase the reliability of a classifier. For instance, Fast Gradient Sign Method (FGSM) \cite{goodfellow2014explaining}, Fast Gradient Sign Method (FGSM) and its variants \cite{kurakin2016adversarial}, and Projected Gradient Descent (PGD) \cite{madry2017towards} use the Saliency gradient to evaluate how much an input sample must be modified to cheat the classifier. However, we want to extend these works, in order to evaluate how reliable is a classifier. We do not focus on how much an image has to be modified, but to see if the generated image can also cheat a human eye. 

To answer this question, we have conducted an experiment, that is shown in Table \ref{tab:robustness}. Given an original image and a trained classifier, we use the saliency output to make perturbations in this image, in order to cheat the classifier and obtain wrong predictions with a high confidence (higher than $95\%$). The first row shows the perturbations needed when using a WRN-16-4 model, which achieves a $99.69\%$ accuracy on the testing set. We can see that the resulting images have no substantial visual difference with respect to the original one. It looks like only random noise was added to the original image. To put it in different words, ten different images that look extremely similar are able to achieve completely different predictions in our WRN-16-4 model. 

As we conclude that introducing white noise can easily fool our WRN-16-4 model, we re-trained it again, but this time introducing some random Gaussian noise in the training images. Apparently, the quality of the classifier decreases, as we only obtain a $99.37\%$ accuracy on the testing set. However,when we take a look at the images (Table \ref{tab:robustness}, second row), we can see that the perturbations look completely different from the original image, and that it should be easy to establish the classifier prediction by just looking at the input. Therefore, we conclude that, although the second model achieves a lower accuracy in the training set, it is a more \textit{reliable model}.

This is a very interesting effect that could led to potential implications in the training algorithm. As we can easily obtain images that are able to \textit{fool} our model, we can make adjustments to our training set so we can improve the \textit{reliability} of our model. This is a very important factor in topics like decision support systems, as we can base our conclusion in a more solid explanation.

It could also have potential implications in medical analysis, as, having any given sample, we can show to doctors which is our conclusion, and how the input should be modified to modify the prediction. Together with a dictionary of potential treatments and their effect to our input parameters, it could also led to treatment recommendations.

\section{Conclusion}
\label{sec:conclusion}

In this paper we have presented a novel Feature Selection ranking approach, called Saliency-based Feature Selection (SFS). Contrary to classic Feature Selection approaches, our algorithm is able to rank the importance of each feature at an instance-level, rather than as a whole dataset. Besides this advantage, experimental results in challenging datasets show that our algorithm is able to achieve state-of-the-art results under different configurations, making it suitable to be used in any king of problem, either is it classification or regression. Contrary to classic Information-based Feature Selection techniques, the reduced complexity of our SFS algorithm (it can be computed simultaneously with the classification or regression training) allows it to be used in high-dimension datasets.

As future research, we aim to use our SFS technique to define a metric that evaluates a model's \textit{degree of robustness}, that is, to measure how hard is to fool either a classifier or a regression architecture, instead of just using visual cues. We also want to test our adversarial images as part of a explainable model in a real scenario like medical information.

\section{Acknowledgements}
This research has been financially supported in part by the Spanish Ministerio de Econom\'ia y Competitividad (research project TIN2015-65069-C2-1-R), and by the Xunta de Galicia (research projects ED431C 2018/34 and Centro singular de investigaci\'on de Galicia, accreditation 2016-2019), and the European Union (European Regional Development Fund - ERDF). We also gratefully acknowledge the support of NVIDIA Corporation with the donation of the Titan Xp GPU used for this research. Brais Cancela acknowledges the support of the Xunta de Galicia under its postdoctoral program.

\section*{References}

\bibliographystyle{unsrt}

\bibliography{egbib}

\end{document}